\documentclass[10pt,twocolumn,letterpaper]{article}

\usepackage[pagenumbers]{cvpr}

\usepackage{amsmath}
\usepackage{multirow}
\usepackage{caption}
\usepackage{url}
\usepackage{subcaption}
\usepackage{wrapfig}
\usepackage{bm}
\usepackage{anyfontsize}
\usepackage[ruled,vlined]{algorithm2e}
\usepackage[dvipsnames]{xcolor}
\usepackage[accsupp]{axessibility}  

\newlength\savewidth\newcommand\shline{\noalign{\global\savewidth\arrayrulewidth
		\global\arrayrulewidth 1pt}\hline\noalign{\global\arrayrulewidth\savewidth}}

\usepackage{diagbox}

\definecolor{cvprblue}{rgb}{0.4,0.15,0.95}
\usepackage[pagebackref=true,breaklinks=true,colorlinks=true,bookmarks=false]{hyperref}
\definecolor{deepred}{HTML}{940000}
\hypersetup{linkcolor=[rgb]{0.4,0.15,0.95}}
\hypersetup{urlcolor  = [rgb]{0.4,0.15,0.95}}
\hypersetup{citecolor=[rgb]{0.4,0.15,0.95}}

\usepackage{colortbl}
\definecolor{Gray}{gray}{0.94}
\definecolor{darkspringgreen}{rgb}{0.09, 0.45, 0.27}

\usepackage{pifont}
\usepackage{tocloft}
\usepackage[toc,page,header]{appendix}
\usepackage{adjustbox}
\usepackage{minitoc}

\renewcommand \thepart{}
\renewcommand \partname{}

\title{\vspace{-8.5mm}\name: Explainable Video Anomaly Detection via Verbalized Learning of Vision-Language Models}

\newcommand{\name}{VERA}

\vspace{-5.5mm}

\author{\fontsize{11.5pt}{\baselineskip}\selectfont
	Muchao Ye\textsuperscript{1*} 
	~~~~~Weiyang Liu\textsuperscript{2}~~~~~Pan He\textsuperscript{3}\\[0.5mm]\fontsize{11pt}{\baselineskip}\selectfont
	\textsuperscript{1}The University of Iowa~~~\textsuperscript{2}Max Planck Institute for Intelligent Systems, T\"ubingen~~~\textsuperscript{3}Auburn University\\\normalsize{
		$^1${\tt muye@uiowa.edu} \ $^2${\tt weiyang.liu@tuebingen.mpg.de} \ $^3${\tt pan.he@auburn.edu}
	} $^*${Corresponding Author} \\
	\normalsize{\url{https://vera-framework.github.io}}
}

\begin{document}
	\maketitle
	\doparttoc 
	\faketableofcontents
	
	\begin{abstract}
		\vspace{-5.5mm}

		The rapid advancement of vision-language models (VLMs) has established a new paradigm in video anomaly detection (VAD): leveraging VLMs to simultaneously detect anomalies and provide comprehendible explanations for the decisions. Existing work in this direction often assumes the complex reasoning required for VAD exceeds the capabilities of pretrained VLMs. Consequently, these approaches either incorporate specialized reasoning modules during inference or rely on instruction tuning datasets through additional training to adapt VLMs for VAD. However, such strategies often incur substantial computational costs or data annotation overhead. To address these challenges in explainable VAD, we introduce a verbalized learning framework named VERA that enables VLMs to perform  VAD without model parameter modifications. Specifically, VERA automatically decomposes the complex reasoning required for VAD into reflections on simpler, more focused guiding questions capturing distinct abnormal patterns. It treats these reflective questions as learnable parameters and optimizes them through data-driven verbal interactions between learner and optimizer VLMs, using coarsely labeled training data. During inference, VERA embeds the learned questions into model prompts to guide VLMs in generating segment-level anomaly scores, which are then refined into frame-level scores via the fusion of scene and temporal contexts. Experimental results on challenging benchmarks demonstrate that the learned questions of VERA are highly adaptable, significantly improving both detection performance and explainability of VLMs for VAD.

	\end{abstract}   
	\vspace{-1mm}
	\section{Introduction}
	\label{sec:intro}

	\begin{figure}[t]
		\centering
		
		\vspace{-1.75mm}
		\includegraphics[width=1\linewidth]{ 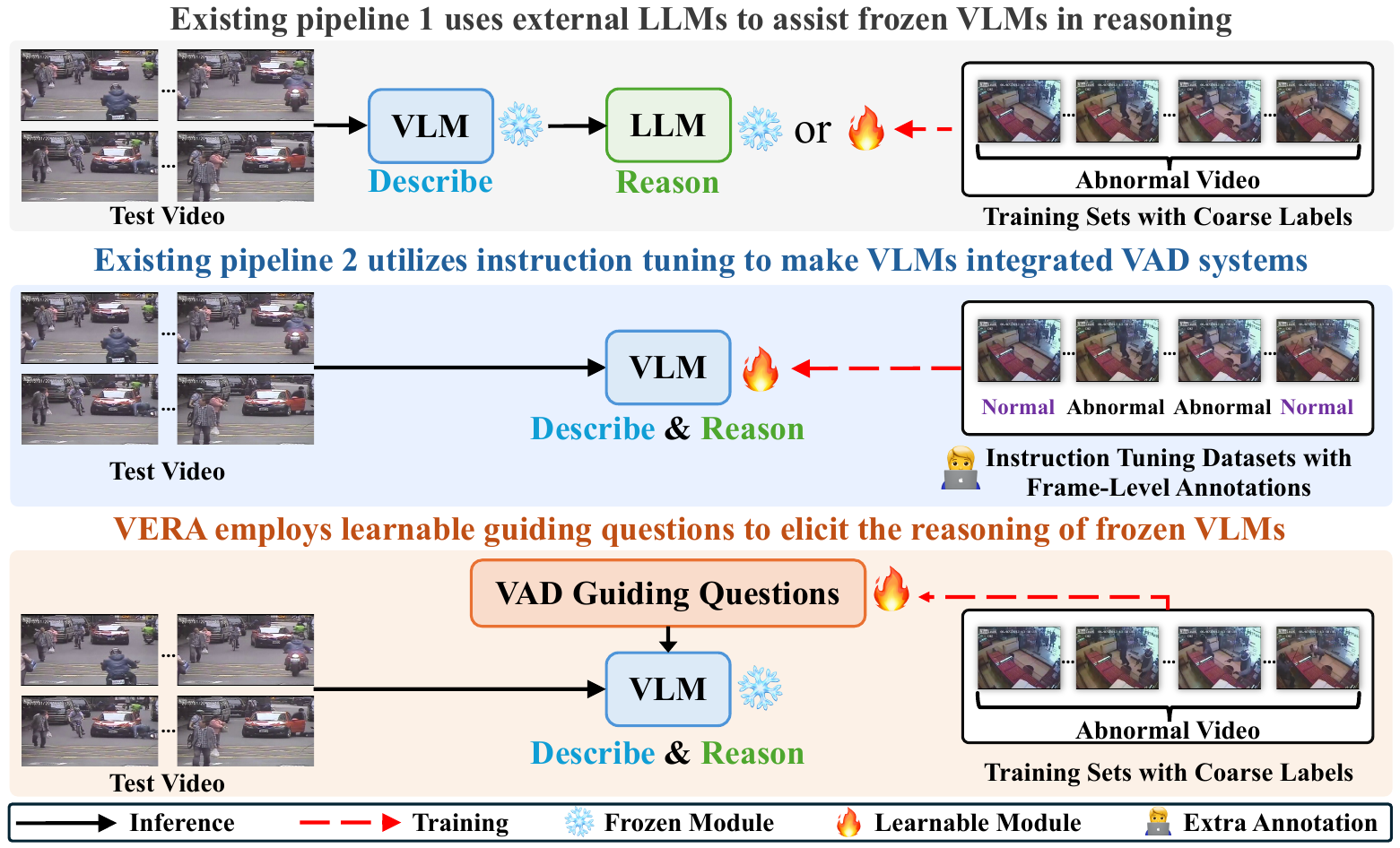}
		\vspace{-6.5mm}
		\caption{\footnotesize {\name} renders frozen VLMs to describe and reason with learnable guiding questions learned from coarsely labeled data.}
		\label{fig:intro}
		\vspace{-3.75mm}
	\end{figure}

	Video anomaly detection~(VAD) aims to automatically identify unexpected and abnormal events in video sequences, with broad applications ranging from autonomous driving~\cite{bogdoll2022anomaly} to industrial manufacturing~\cite{roth2022towards}. While achieving good performance in VAD is essential, providing clear explanations for detected anomalies is even more crucial. 
	
	To this end, our work primarily focuses on explainable VAD, which requires both comprehensive visual understanding and the ability to generate human-interpretable predictions. The rapid advancement of vision-language models (VLMs)~\cite{li2023blip,chen2024internvl,zhuminigpt,liu2024visual} enables us to address both requirements through their strong visual reasoning and language interaction capabilities.  As multi-modal architectures that effectively combine the reasoning capabilities from large language models (LLMs)~\cite{DBLP:conf/nips/BrownMRSKDNSSAA20} and the visual understanding capabilities from pretrained vision encoders~\cite{DBLP:conf/iclr/DosovitskiyB0WZ21}, VLMs are particularly well-suited for VAD for they can offer explainable predictions that clearly illustrate the rationale behind specific anomalies, making the results more interpretable to users. Recent research on VAD has consequently focused on how to effectively leverage the power of pretrained VLM. As shown in Fig.~\ref{fig:intro}, existing approaches aim to address the misalignment problem between VLMs' pretraining tasks and the VAD requirements through either  additional reasoning modules or instruction tuning (IT):

	\begin{itemize}
		\item One line of research \emph{introduces external LLMs to assist frozen VLMs to reason in VAD}~\cite{zanella2024harnessing,yang2024follow}.  
		It uses VLMs to caption what they see given a video, and the descriptions are then passed to an external LLM, \eg, GPT-4~\cite{achiam2023gpt}, to reason whether an anomaly occurs. 
		
		\item Another line of research, instead, \emph{expands VLMs to generate explainable prediction via IT}~\cite{lv2024video,zhang2024holmes}. 
		This research line creates additional VAD datasets with frame-level annotations and leverages exemplary instructions to fine-tune the VLM, enabling it to detect anomalies and generate human-interpretable explanations.

	\end{itemize}

	\noindent \textbf{Key Observations and Research Question}. While prior research demonstrates the potential of applying VLMs to VAD, we identify that this new paradigm is hindered by a shared critical issue: the use of additional reasoning modules or fine-grained labeled datasets incurs significant computational cost either in the inference or training phases. First, decoupling a VAD system into a frozen VLM and an extra LLM introduces more overhead in inference, because it separates the description generation and reasoning processes. Secondly, although IT-based methods enable VLMs to effectively integrate description and reasoning for VAD, they require additional manpower and computational resources for annotating and finetuning on fine-grained labeled instruction datasets, which is time-consuming and not scalable for large-scale datasets. In light of this, we investigate the following unexplored yet important question:
	
	\vspace{2mm}
	\emph{Can we enable a frozen VLM to integrate description and reasoning for VAD without instruction tuning?}
	
	\vspace{2mm}
	\noindent \textbf{Our Approach}.  This research question is nontrivial because the reasoning ability of a frozen VLM is limited in general visual tasks, and it struggles to handle complex reasoning tasks like VAD, which requires the understanding of subtle, context-dependent outliers. To illustrate, Table~\ref{table:intro} shows that prompting frozen VLMs with simple VAD questions used in existing works leads to unsatisfactory results. Thus, instruction-tuning a VLM seems necessary to make it responsive to specific instructional cues and capture delicate visual variations. In this paper, we question the necessity of such an operation and propose a principled approach to tailor frozen VLMs for VAD. 
	
	Specifically, our solution is guided by the intuition that the reasoning ability of VLMs for VAD will improve if we find questions with suitable and concrete description of abnormal patterns rather than with abstract and general words like ``anomaly'' to prompt them. Our idea is to iteratively refine anomaly descriptions from abstract ones (\eg, ``is there any anomaly?'') to detailed, specific characterizations.
	
	Driven by such insight, we propose a framework, termed {\name}, to explore \underline{ver}balized learning (VL) for V\underline{A}D. This framework considers the practical constraint that it is suboptimal to manually write down VAD guiding questions across VLMs, so it introduces a data-driven learning task to identify suitable anomaly-characterization questions containing concrete abnormal patterns for the frozen VLM using coarsely labeled datasets, eliminating the need for IT. Specifically, in the \underline{training phase}, VERA treats the questions guiding the reasoning of VLMs in VAD as learnable parameters, improving them based on the verbal feedback from an optimizer VLM on the performance of a learner VLM on an intermediate VAD subtask---binary video classification for each video in the VAD training set. This design is both efficient and appropriate for VAD, as it accounts for video-specific properties like temporality while relying solely on provided coarse video-level labels.
	After that, considering the large scale of video frames, {\name} assigns a fine-grained anomaly score for each frame in a coarse-to-fine manner in the \underline{inference phase}. First, {\name} generates segment-level anomaly scores by querying VLMs with the learned guiding questions. Next, {\name} improves the initial score by incorporating scene context into each segment score via ensembling. Finally, {\name} outputs frame-level scores by fusing temporal context via Gaussian smoothing and frame-level position weighting.

	\begin{table}[t]
		\centering
		
		\small
		\setlength{\tabcolsep}{1pt}
		\renewcommand{\arraystretch}{1.1}
		\hspace*{-0.2cm}
		\begin{tabular}{c|c}
			VAD Question for InternVL2-8B & AUC (\%) \\
			\shline
			``Describe the video and is there any anomaly?''~\cite{lv2024video} & 53.05 \\
			``Are there any abnormal events in the video?''~\cite{zhang2024holmes} &  65.03  \\
		\end{tabular}
		\vspace{-2.5mm}
		\caption{\footnotesize Instructing a frozen VLM (InternVL2-8B~\cite{chen2024internvl}) with simple questions to perform VAD yields poor AUC on UCF-Crime~\cite{sultani2018real} dataset.}
		\label{table:intro}
		\vspace{-4mm}
	\end{table}
	
	\vspace{0.5mm}
	\noindent \textbf{Contributions}. To sum up, our contributions are: 
	
	\begin{itemize}
		\item To our knowledge, we present the first approach, that is, {\name}, to adapt frozen VLMs as an integrated system for VAD by learning detailed anomaly-characterization questions in prompts that decompose anomalies into concrete and recognizable patterns. {\name} learns them directly from coarsely labeled datasets, eliminating the need for IT or external reasoning modules.
		\item We introduce an effective VL-based algorithm for VLMs in VAD, allowing direct adaptation without modifying model parameters. With coarse labeled VAD datasets only, our approach obtains good guiding questions in VAD by relying on the verbal interaction between learner and optimizer VLMs in verbalized training. Additionally, we design a coarse-to-fine strategy to derive frame-level anomaly scores from verbally learned guiding questions in VAD, integrating both scene and temporal contexts for better VAD performance and reasoning.
		\item  The learned guiding questions from {\name} are expressed in natural languages, providing a unified method to encode and transfer prior VAD knowledge seamlessly to other datasets or VLMs. In challenging VAD datasets like UCF-Crime~\cite{sultani2018real} and XD-Violence~\cite{wu2020not}, {\name} achieves state-of-the-art explainable VAD performance and enjoys good generalization ability across models and datasets.

	\end{itemize}

	\section{Related Work}
	\noindent\textbf{Video Anomaly Detection}. VAD is the task of localizing frames that contain abnormal events in a given video. This task is challenging for anomalies cover a broad scope of events like accidents and criminal activities while training sets only offer coarse annotations. Modern VAD methods are based on deep neural networks (DNNs) for their superiority and are going through a paradigm shift in using VLMs: (1) Early DNNs for VAD are task-specific, which often employ unsupervised (including one-class) or weakly supervised (WS) learning techniques for training.  Most unsupervised learning methods~\cite{liu2018future,ye2019anopcn,zhang2024multi,wang2019gods,lu2013abnormal,tur2023unsupervised} train DNNs on frame reconstruction/prediction tasks to establish representation spaces for normal/abnormal videos. WS learning methods~\cite{sultani2018real, chen2024prompt, yang2024text, zhang2023exploiting, lv2023unbiased,wu2024vadclip} leverage both normal and abnormal videos to train a feature extractor that distinguishes anomalies from normalcy, typically using multiple instance learning~\cite{sultani2018real} objectives. (2) Recent VAD methods adopt VLMs due to their remarkable success across core vision tasks~\cite{pratt2023does,liu2024visual,guo2024regiongpt, tang2024hawk}. Early research~\cite{zhang2024holmes, zanella2024harnessing, yang2024follow, lv2024video} has leveraged VLMs to generate textual descriptions of detected anomalies to enhance prediction explainability for VAD. However, current approaches incur high processing demands from external LLMs or require substantial effort and cost for fine-tuning on additional datasets, which are computationally inefficient in training or inference. Our work reduces the processing overhead by adapting frozen VLMs for VAD without model parameter modification or extra reasoning modules via learnable guiding questions, which elicit superior reasoning from frozen VLMs and significantly boost their performance in VAD.

	\vspace{0.5mm}
	\noindent \textbf{Verbalized Learning for VLMs}. The designed VL framework is inspired by a recent technique called verbalized machine learning (VML)~\cite{xiao2024verbalized}. The main idea of VML is to use LLMs to approximate functions and learn the verbal rules and descriptions of performing specific tasks, which casts traditional machine learning tasks as language-based learning tasks. This approach regards the language expressions that define classification rules and other task-specific criteria as learned parameters, and optimize them in a data-driven fashion through interactions between a learner and an optimizer modeled by LLMs or VLMs. However, the VML framework is limited to tasks involving regression on scalar values or classification for static images. 
	A similar idea has also been explored in a concurrent method, TextGrad~\cite{yuksekgonul2024textgrad}, which integrates the process of incorporating textual feedback from LLMs for improving prompts in PyTorch and further proves its effectiveness in coding, question answering, and optimization in chemistry and medicine.
	Compared to existing works, our work pioneers VL for the VAD task and video data, which remains unsolved for previous VL frameworks focus on static-data tasks and cannot handle the challenges of temporality and scene dynamics in videos. Specifically, {\name} introduces a new learning paradigm for VAD: generating effective questions that encapsulate key abnormal patterns in videos to elicit the reasoning ability from VLMs for explainable VAD. Additionally, {\name} works for any VAD dataset and supports WS learning. Unlike previous WS methods, {\name} only needs to learn concise text but not millions of parameters, so the training is lightweight.

	\section{The VERA Framework}

	\begin{figure*}
		\centering
		
		\includegraphics[width=1\linewidth]{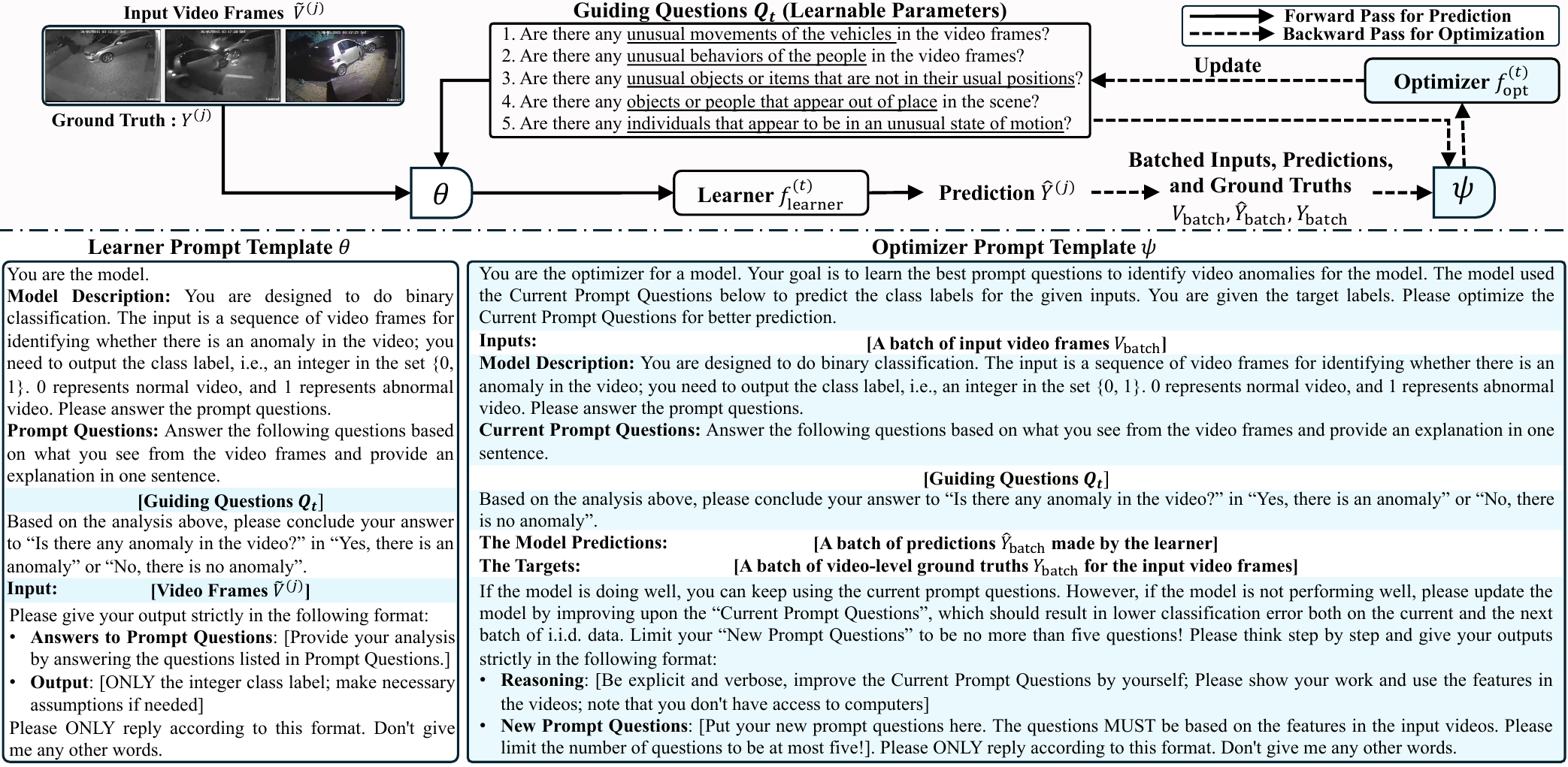}
		\vspace{-6mm}
		\caption{\footnotesize The overall training pipeline in {\name} aims to optimize VAD guiding questions iteratively.  In each iteration $t$, the optimization is verbalized by providing verbal instructions for the learner and optimizer to follow. They will generate predictions and new guiding questions, respectively.}
		\label{fig:qvml}
		\vspace{-3mm}
	\end{figure*}

	Our approach adapts VLMs to detect video anomalies without additional reasoning modules or IT. We now formulate the VAD task and detail the design of {\name}.

	\subsection{Problem Formulation}
	
	\noindent \textbf{Video Anomaly Detection}. Let \( V \) be a video with \( F \) frames, represented as \( V = \{I_i\}_{i=1}^F \), where \( I_i \) is the $i$-th frame \( (1 \le i \le F) \). Our objective is to locate and detect the start and end of anomalous events within \( V \). In standard labeling, any frame associated with an anomaly is labeled as 1, and normal frames are labeled as 0. Therefore, the ground truth label sequence for \( V \) is \( Y = [y_1, \dots, y_F] \), where \( y_i \in \{0, 1\} \) represents the fine-grained label for \( I_i \). We aim to use a frozen VLM, \( f_{\text{VLM}} \), to generate anomaly score predictions across all frames, \( \hat{Y} = [\hat{y}_1, \dots, \hat{y}_F] \), where \( \hat{y}_i \in [0, 1] \) is a continuous anomaly score for $I_i$.
	
	\vspace{0.5mm}
	\noindent \textbf{Available Training Data for VAD}. Typically, VAD datasets only provide coarsely labeled training sets~\cite{sultani2018real,wu2020not,liu2018future,lu2013abnormal}. We denote a VAD training set as \( \mathcal{D} = \{(V^{(j)}, Y^{(j)})\}_{j=1}^{N} \), where \( N \) is the total number of training videos, \( V^{(j)} \) represents the \( j \)-th video \( (1\le j\le N) \)  and \( Y^{(j)} \) is the corresponding video-level label.  \( Y^{(j)} = 1 \) if $V^{(j)}$ contains any anomaly defined by the dataset annotators, \eg, abuse or arson activities, and \( Y^{(j)} = 0 \) if \( V^{(j)} \) has no anomalies. For \( V^{(j)} \), we suppose it contains \( F_j \) frames and denote the frames sequence as \( V^{(j)} = \{ I_i^{(j)} \}_{i=1}^{F_j} \), where \( I_i^{(j)} \) is the \( i \)-th frame (\( 1 \le i \le F_j \)) in $V^{(j)}$.

	\subsection{Training in {\name}} 
	\label{sec:train}

	\noindent \textbf{Training Objective}. We aim to learn guiding questions that break down a complex and ambiguous concept (\ie, what is an ``anomaly'') into a set of identifiable anomalous patterns to unlock reasoning capabilities within frozen VLMs for VAD tasks. Those patterns vary among datasets, making manually designed descriptions ineffective for generalization. To address this, we propose a general VL framework shown in Fig.~\ref{fig:qvml} to generate the desired guiding questions. We denote the guiding question set as $\mathbf{Q} = \{q_1, \dots, q_m\}$, where $q_i$ is the $i$-th question ($1\le i \le m$) and $m$ is the number of questions. The training framework considers  $\mathbf{Q}$ as the \underline{learnable parameters},  which are optimized through verbal interaction between a learner and an optimizer, modeled by VLMs through leveraging their ability to follow instructions with given prompts.

	\noindent \textbf{Training Data}. The training data for learning $\mathbf{Q}$ consist of paired sampled video frames and video-level labels.  Sampling is necessary because the amount of video frames is so huge that we cannot compute with every frame. We explore three types of sampling strategies and find that uniform sampling~\cite{zhang2023video} yields the best results. That is, with any video $V^{(j)} \in \mathcal{D}$, we first calculate the interval between sampled frames as $l = \text{floor}(F_j / S)$, where $S$ is the number of sampled frames, and $\text{floor}$ denotes rounding down to the nearest integer. Given $l$, the uniformly sampled frames from $V^{(j)}$ are represented by $\tilde{V}^{(j)} = [I_{1}^{(j)}, I_{l+1}^{(j)},  \dots, I_{(S-1)\cdot l+1}^{(j)}]$. The label used for training is $Y^{(j)}$ only, resulting in training data pairs $\{(\tilde{V}^{(j)}, Y^{(j)})\}_{j=1}^{N}$ for {\name}.
	
	\vspace{0.5mm}
	\noindent \textbf{Updating $\mathbf{Q}$ via Learner and Optimizer}. Since $\mathbf{Q}$ are verbal expressions for specific anomaly patterns, {\name} inherits the idea of VML~\cite{xiao2024verbalized} in training:  optimizing language-based parameters by verbal communication between a learner agent $f_{\rm learner}$ and an optimizer agent $f_{\rm opt}$, rather than by numerical optimization algorithms like Adam~\cite{kingma2014adam}. W.l.o.g., we take an arbitrary iteration $t$ when implementing the complete algorithm (detailed in Supplementary Material) for illustration. We denote any LLM-based model as $f(x; \phi)$ where $x$ represents the input data, and $\phi$ denotes the natural language instructions for $f$ to follow, which is considered as learnable parameters in our VL framework. Specifically, $\mathbf{Q}$ contains parameters to be learned in {\name}. As depicted in Fig.~\ref{fig:qvml}, in each iteration $t$, the learner agent $f_{\text{learner}}^{(t)}$ is modeled by the frozen VLM $f_{\text{VLM}}(\cdot)$  used for VAD  with a specific prompt template $\theta$ that guide $f_{\text{VLM}}(\cdot)$ to conduct a learning task by pondering on current guiding questions $\mathbf{Q}_t$. We denote the learner agent as 
	$f_{\text{learner}}^{(t)}(x) = f_{\text{VLM}}(x; (\theta,\mathbf{Q}_t))$, where $x$ is the input in a learning task, and $\mathbf{Q}_t$, the learnable guiding questions applied in each iteration $t$, constitutes the core parameters that distinguish the learner between iterations. Meanwhile, we introduce an optimizer $f_{\text{opt}}^{(t)}$ to assess the quality of the predictions of the learner and to optimize $\mathbf{Q}_t$. W.l.o.g., we use the same frozen VLM $f_{\rm VLM}$ to model the optimizer. As demonstrated in Fig.~\ref{fig:qvml}, we provide another specific prompt template $\psi$ for the learner to follow to optimize $\mathbf{Q}_{t}$, so we denote the optimizer agent as $f_{\text{opt}}^{(t)}(z) = f_{\text{VLM}}(z; (\psi, \mathbf{Q}_t))$, where $z$ is its input and $\psi$ is the instruction to improve $\mathbf{Q}_t$.
	It is important to note that $ f_{\text{learner}}^{(t)} \neq f_{\text{opt}}^{(t)}$ because $f_{\text{learner}}^{(t)}$ follows $(\theta,\mathbf{Q}_t)$ to conduct a learning task, while $f_{\text{opt}}^{(t)}$ follows $(\psi,\mathbf{Q}_t)$ to refine $\mathbf{Q}_{t}$.
	
	\underline{\emph{Learning Task for $f_{\rm learner}$}}. The learner executes the ``forward pass'' and outputs a prediction. Recall that we only use the original coarsely labeled information for training. Thus, we design a binary classification task for $f_{\rm learner}$, which accounts for the temporal nature of video data, the sparsity of anomalies, and the weak supervision in VAD datasets. In this task, the job of the learner $f_{\rm learner}$ is to produce a binary classification prediction $\hat{Y}^{(j)}$ to determine whether there is an anomaly in the video based on the sampled frames $\tilde{V}^{(j)}$. As shown in Fig.~\ref{fig:qvml}, we explain the task in natural language in the ``Model Description'' section in $\theta$. Guiding questions $\mathbf{Q}_t$ are inserted in the ``Prompt Questions'' section in $\theta$ to elicit reasoning of the VLM. This template design is based on the prompt structures used in VML, with targeted modifications to help the learner effectively address this WS learning task. Given $\theta$ and a sampled frame set $\tilde{V}^{(j)}$, the learner will output a prediction as
	\begin{equation}
		\hat{Y}^{(j)} = f_{\rm learner}^{(t)}(\tilde{V}^{(j)}),
		\label{eq:predict}
	\end{equation}
	where $\hat{Y}^{(j)}=1$ if the learner thinks there is an anomaly after skimming across the sampled frames $\tilde{V}^{(j)}$ and reasoning through the guiding questions $\mathbf{Q}_t$, and otherwise, $\hat{Y}_i=0$.

	\underline{\emph{Optimization Step in $f_{\rm opt}$}}. The optimizer executes the ``backward pass'' to update the questions $\mathbf{Q}_{t}$ via a mini-batch (batch size is $n$). Suppose the visual input in a batch is $V_{\rm batch}=[\tilde{V}^{(1)}_{\rm batch}, \cdots, \tilde{V}^{(n)}_{\rm batch}]$ and the corresponding ground truths are $Y_{\rm batch}=[Y^{(1)}_{\rm batch},\cdots,Y^{(n)}_{\rm batch}]$.  The learner generates prediction as $\hat{Y}_{\rm batch}=[\hat{Y}^{(1)}_{\rm batch}, \cdots, \hat{Y}^{(n)}_{\rm batch}]$ with the current questions $\mathbf{Q}_t$ by Eq.~\eqref{eq:predict}. The optimizer will output a new set of questions $\mathbf{Q}_{t+1}$ by following the prompt $\psi$ with batched data.  We denote the optimization step as
	\begin{equation}
		\mathbf{Q}_{t+1} = f_{\rm opt}^{(t)}(V_{\rm batch}, \hat{Y}_{\rm batch}, Y_{\rm batch}),
		\label{eq:optimize}
	\end{equation}
	where $\mathbf{Q}_{t+1}$ is a new set of guiding questions constructed from $f_{\rm opt}^{(t)}$ owing to its text generation and instruction following abilities after reading $\psi$.

	\vspace{0.5mm}
	
	\subsection{Inference in {\name}}
	\label{sec:inference}
\begin{figure}[t]
		\centering
		\includegraphics[width=\linewidth]{ 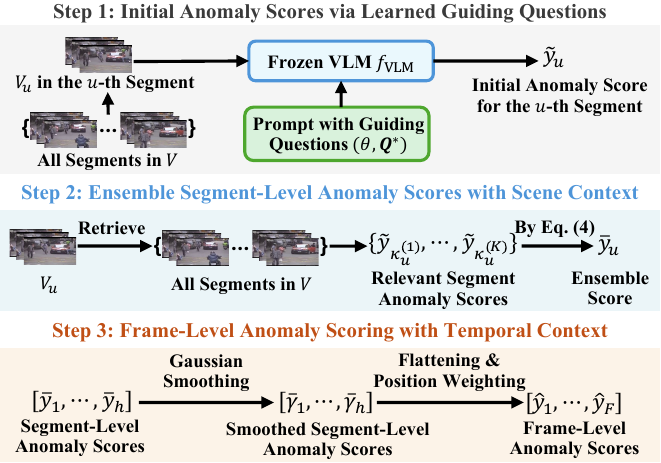}
		\vspace{-6.7mm}
		\caption{\footnotesize VERA computes anomaly scores with $\mathbf{Q}^{*}$ in three steps. 
		}
		\label{fig:score}
		\vspace{-4.7mm}
	\end{figure}

	During training, we denote the one with the largest validation accuracy as $\mathbf{Q^{*}}$. In inference, given $\mathbf{Q}^{*}$, {\name} yields fine-grained anomaly score $\hat{Y}$ for a test video $V$ via a coarse-to-fine process shown in Fig.~\ref{fig:score}.

	\noindent\textbf{Step 1: Initial Anomaly Scores via Learned Guiding Questions}. We divide the video into segments and analyze each segment independently first. Following~\cite{zanella2024harnessing}, we perform equidistant frame sampling within $V$ to obtain the set of segment centers $\mathcal{C}= \{I_{1},  I_{d+1},\cdots,I_{(h-1)\cdot d +1}\}$, where $d$ is the interval between centers and $h={\rm floor}(F/d)$ is the total number of segments. For each center frame $I_{(u-1) \cdot d + 1}$ ($1 \leq u \leq h$), we define a 10-second window around it as the $u$-th segment, within which we uniformly sample 8 frames. We denote the sampled frame set in the $u$-th segment as $V_{u}$. Next, we input $V_{u}$ in $f_{\rm VLM}$ with the prompt $(\theta, \mathbf{Q}^*)$ to get the initial score
	\begin{equation}
		\tilde{y}_{u} = f_{\rm VLM}(V_{u}; (\theta ,\mathbf{Q}^{*})),
		\label{eq:initial}
	\end{equation}
	where $\tilde{y}_{u}=1$ if $f_{\rm VLM}$ thinks the segment contains an anomaly  after reasoning via $\mathbf{Q}^{*}$ with $V_{u}$, and otherwise, $\tilde{y}_{u}=0$. By repeating Eq.~\eqref{eq:initial} for each segment, we have a segment-level initial anomaly score set $\tilde{Y} = [\tilde{y}_{1},\cdots,\tilde{y}_{h}]$. 
	
	\vspace{0.5mm}
	\noindent\textbf{Step 2: Ensemble Segment-Level Anomaly Scores with Scene Context}. Note that the scores derived above only examine a short moment in a long video without considering any context. To resolve it, we refine the initial segment-level score by incorporating scene context---defined as preceding and following segments that contain similar elements, such as actors and background, to those in the current segment.

	We measure the relevance between different video segments by the cosine similarity of their feature representations~\cite{liu2017deep}, extracted by a pretrained vision feature extractor $g$, \eg, ImageBind~\cite{girdhar2023imagebind}. For the $u$-th segment $V_{u}$, its similarity with any segment $V_{w}$ ($1\le w\le h)$ is ${\rm sim}(u, w) = {\rm cos}\left(\frac{e_{u} \cdot e_{w}}{||e_{u}|| \cdot ||e_{w}||}\right)$, where ${\rm cos}$ denotes the cosine function, and $e_{u} = g(V_u)$ and $e_{w} = g(V_w)$ represent their features. Let $\kappa_{u} = [\kappa_{u}^{(1)}, \dots, \kappa_{u}^{(K)}]$ denote the indices of the top-$K$ segments similar to $V_u$. We refine the anomaly score by
	
	\vspace{-5.8mm}
	\begin{equation}
		\bar{y}_{u} = \sum_{i=1}^{K} \tilde{y}_{\kappa_{u}^{(i)}}\cdot\frac{{\rm exp}({\rm sim}(u,\kappa_{u}^{(i)})/\tau)}{\sum_{j=1}^{K} {\rm exp}( {\rm sim}(u,\kappa_{u}^{(j)})/\tau )},
		\label{eq:weight}
	\end{equation}
	where $\bar{y}_{u}$ is an ensemble of initial scores of top-$K$ video segments relevant to $V_{u}$. Here, the initial score of each retrieved segment is weighted by a factor derived from the cosine similarity and normalized by the Softmax function (with $\tau$ as the temperature hyperparameter). Accordingly,  scenes with greater similarity are assigned higher weights, making the ensemble score a more comprehensive reflection of anomalies with the video context. By applying Eq.~\eqref{eq:weight} for all segments, we obtain 
	$\bar{Y} = [\bar{y}_{1}, \dots, \bar{y}_{h}]$.
	
	\vspace{0.5mm}
	\noindent\textbf{Step 3: Frame-level Anomaly Scoring with Temporal Context}. Given $\bar{Y}$, we aim to incorporate temporal context to capture how events evolve over time when computing frame-level anomaly scores, for the abnormality of an event often depends on the timing and progression of observed activities. To detail, we first apply Gaussian smoothing~\cite{gonzalez2009digital} to aggregate local temporal context into the segment-level anomaly scores. We denote the Gaussian kernel (suppose the filter size is $\omega$) as $G(p)={\rm exp}(\frac{-p^2}{2\sigma_{1}^2})$ where $p$ is the distance from the kernel center and $\sigma_{1}$ is the variance. We update segment-level scores as $\bar{\Gamma}=\bar{Y}*G=[\bar{\gamma}_1,\cdots,\bar{\gamma}_{h}]$, where $*$ is the convolution operation. Next, we integrate global temporal context by position weighting. With $\bar{\Gamma}$, we flatten it into frame-level scores by assigning  the score $\bar{\gamma}_{u}$ to each frame in the $u$-th segment, \ie, $[I_{(u-1)\cdot d+1},\cdots, I_{u\cdot d}]$. We denote the frame-level score sequence after flattening as $[\rho_{1},\cdots,\rho_{F}]$. We then apply the Gaussian function to encode position weights as $w(i) = \exp\left(\frac{-(i - c)^2}{2\sigma_2^2}\right)$, where $i$ $(1\le i\le F)$ is any frame index, $c = \text{floor}(F/2)$ is the center frame index, and $\sigma_2$ is the variance. The anomaly score for the $i$-th frame is:
	\begin{equation}\label{eq:final}
		\hat{y}_{i} = w(i) \cdot \rho_{i}.
	\end{equation}
	This operation scales the score $\rho_{i}$, diminishing the anomaly score for frames near the beginning and end of the event. This helps better capture the temporal progression of anomalies: the score gradually increases as the anomaly reaches its peak and decreases afterward. The final scores is denoted as $\hat{Y} = [\hat{y}_{1}, \dots, \hat{y}_{F}]$ after applying Eq.~\eqref{eq:final}.

	\noindent\textbf{Explainable VAD by VERA}. When using template $\theta$ embedded with $\mathbf{Q}^{*}$ to compute $\hat{Y}$, we ask the VLM to ``provide an explanation in one sentence'' when reasoning, and VLM will explain the anomaly score it assigns based on $\mathbf{Q}^{*}$.

	\section{Experiments and Results}
	In this section, we present an evaluation of VERA as follows, addressing key questions of interest including: (Q1) Does it enhance the effectiveness of frozen VLMs in VAD? (Q2) Is its design reasonable and well-structured? (Q3) How well does it generalize across different scenarios?  
	
	\subsection{Experimental Settings}
	\label{sec:exp_setting}

	\noindent \textbf{Datasets}. We conduct experiments on two large-scale VAD datasets: (1) UCF-Crime~\cite{sultani2018real}  collected from surveillance videos with 13 types of anomalies and 290 (140 abnormal) test videos (2.13 minutes long on average). (2) XD-Violence~\cite{wu2020not} with 6 anomaly categories and 800 (500 abnormal) test videos (1.62 minutes long on average).

	\noindent \textbf{Metrics}. Following approaches in~\cite{zanella2024harnessing,zhang2024holmes}, we mainly evaluate VAD performance using the Area Under the Curve (AUC) of the frame-level Receiver Operating Characteristic (ROC) curve, as it provides a comprehensive measure of model performance across all thresholds.  
	
	\noindent \textbf{Baselines}. We categorize baselines into non-explainable approaches and explainable ones as~\cite{zhang2024holmes} does. Non-explainable ones are obtained by WS learning~\cite{wu2020not,wu2024open,wu2022self,tian2021weakly,li2022self,chen2023mgfn,joo2023clip,sultani2018real,zaheer2022generative,zhong2019graph,feng2021mist,zaheer2020claws,li2022scale} and  unsupervised learning~\cite{thakare2023dyannet,tur2023unsupervised,thakare2023rareanom, wang2019gods,hasan2016learning,lu2013abnormal}. These non-explainable approaches cannot provide language-based explanations for VAD.  
	For explainable approaches, we use LAVAD~\cite{zanella2024harnessing}, Holmes-VAD~\cite{zhang2024holmes}, and VADor~\cite{lv2024video} as representatives of Pipeline 1 and Pipeline 2 shown in Fig.~\ref{fig:intro}. It should be noted that \cite{yang2024follow} does not report performance on UCF-Crime and XD-Violence. Additionally, we include zero-shot (ZS) VAD by frozen VLMs designed by ~\cite{zanella2024harnessing} as baselines.

	\noindent \textbf{Implementation of {\name}}. In our experiments, we choose a small VLM, InternVL2-8B~\cite{chen2024internvl}, as the backbone $f_{\rm VLM}$ for building~{\name} by default, if not otherwise specified.  We also explore other backbones, such as Qwen2-VL-7B~\cite{Qwen2VL} and InternVL2-40B~\cite{chen2024internvl} for ablation. We train $\mathbf{Q}$ for no more than 10 epochs, with a validation accuracy calculated every 100 iterations to determine $\mathbf{Q}^{*}$. We set $n$ as 2, $S$ as 8, and $m$ as 5 for training. The initial questions $\mathbf{Q}_{0}$ is ``\emph{1. Is there any suspicious person or object that looks unusual in this scene? 2. Is there any behavior that looks unusual in this scene?}'', inspired by previous VAD methods~\cite{wu2024open,hasan2016learning}, which assume anomalies appear with unusual appearance or motions.

	\subsection{Comparison to State-of-the-art Methods}
	
	\setlength{\columnsep}{9pt}
	\begin{wraptable}{r}[0cm]{0pt}
		\hspace{-2.4mm}
		\setlength{\tabcolsep}{4.2pt}
		\renewcommand{\arraystretch}{1.09}
		\footnotesize
		\centering
		
		\begin{tabular}{ l|c}
			\specialrule{0em}{0pt}{-14pt}
			Method & AUC\\
			\shline
			\multicolumn{2}{c}{\emph{Non-explainable VAD Methods}}\\ 
			Wu et al.~\cite{wu2020not}  & 82.44  \\
			OVVAD~\cite{wu2024open} & 86.40  \\
			S3R~\cite{wu2022self} & 85.99   \\
			RTFM~\cite{tian2021weakly} & 84.30   \\
			MSL~\cite{li2022self}  &  85.62  \\
			MGFN~\cite{chen2023mgfn}  & 86.98   \\
			SSRL~\cite{li2022scale} & 87.43\\
			CLIP-TSA~\cite{joo2023clip}  & \textbf{87.58}   \\
			Sultani et al.~\cite{sultani2018real}  & 77.92  \\
			GCL~\cite{zaheer2022generative} & 79.84   \\
			GCN~\cite{zhong2019graph} & 82.12 \\
			MIST~\cite{feng2021mist}  & 82.30   \\
			CLAWS~\cite{zaheer2020claws} & 83.03\\
			DYANNET~\cite{thakare2023dyannet}& 84.50 \\
			Tur el al.~\cite{tur2023unsupervised} & 66.85 \\
			GODS~\cite{wang2019gods} & 70.46   \\
			\shline
			\multicolumn{2}{c}{\emph{Explainable VAD Methods}}\\
			LAVAD~\cite{zanella2024harnessing} & 80.28  \\
			Holmes-VAD~\cite{zhang2024holmes} & 84.61   \\
			VADor~\cite{lv2024video} & 85.90   \\
			ZS CLIP~\cite{zanella2024harnessing} & 53.16  \\
			{\scriptsize ZS IMAGEBIND-I}~\cite{zanella2024harnessing} & 53.65  \\
			{\scriptsize ZS IMAGEBIND-V}~\cite{zanella2024harnessing} & 55.78  \\
			LLAVA-1.5~\cite{liu2024improved} &72.84  \\\rowcolor{Gray}
			{\name} & \textbf{86.55}   \\
			\specialrule{0em}{-7pt}{0pt}
		\end{tabular}
		\caption{\footnotesize AUC (\%) on UCF-Crime. No IT is used for Holmes-VAD and VADor.}
		\label{table:comparison}
		\vspace{-1mm}
	\end{wraptable}

	We address Q1 by empirically comparing {\name} to existing VAD methods. First, in Table~\ref{table:comparison}, {\name} achieves the highest AUC among explainable VAD methods on UCF-Crime, outperforming Holmes-VAD and VADor (without IT, as reported in their papers) in a fair comparison. Importantly, unlike these methods, {\name} does not need to modify the model parameters, demonstrating its suitability to directly adapt VLM to the VAD task with minimal training requirements. Moreover, {\name} surpasses LAVAD by $6\%$ in AUC on UCF-Crime, uniquely integrating both description and reasoning capabilities in VAD. Compared to non-explainable methods, {\name} achieves AUC performance that is comparable to one of the top-performing methods, CLIP-TSA, on UCF-Crime, while offering the additional advantage of explainable predictions. 
	
	\setlength{\columnsep}{10pt}
	\begin{wraptable}{r}[0cm]{0pt}
		\centering
		\footnotesize
		\setlength{\tabcolsep}{6pt}
		\renewcommand{\arraystretch}{1.01}
		\hspace{-2.5mm}
		\begin{tabular}{ l|c}
			\specialrule{0em}{0pt}{-.75pt}
			Method & AUC\\
			\shline
			\multicolumn{2}{c}{\emph{Non-Explainable VAD Methods}}\\
			
			Hasan et al.~\cite{hasan2016learning}  &50.32  \\
			Lu et al.~\cite{lu2013abnormal}  & 53.56  \\
			BODS~\cite{wang2019gods}   &  57.32   \\
			GODS~\cite{wang2019gods}   &  61.56 \\
			RareAnom~\cite{thakare2023rareanom} & \textbf{68.33} \\
			\shline
			\multicolumn{2}{c}{\emph{Explainable VAD Methods}}\\
			LAVAD~\cite{zanella2024harnessing}  & 85.36 \\
			ZS CLIP~\cite{zanella2024harnessing} &  38.21    \\
			{\scriptsize ZS IMAGEBIND-I}~\cite{zanella2024harnessing}  & 58.81  \\
			{\scriptsize ZS IMAGEBIND-V}~\cite{zanella2024harnessing}  & 55.06   \\
			LLAVA-1.5~\cite{liu2024improved}   &  79.62   \\\rowcolor{Gray}
			{\name}  & \textbf{88.26} \\
			\specialrule{0em}{-8.25pt}{0pt}
			
		\end{tabular}
		\caption{\footnotesize AUC (\%) on XD-Violence.}
		\label{table:comparison_xd}
		\vspace{-8mm}
	\end{wraptable}
	
	Similar advantages are also observed in Table~\ref{table:comparison_xd} for XD-Violence. Considering multiple factors, including performance, training efficiency, system integration, and explainability, {\name} stands out as a promising pipeline for  VLMs in VAD.

	\subsection{Ablation Studies}
	\label{sec:ablation}

	We perform necessary ablation studies on UCF-Crime to answer both Q2 and Q3 for a comprehensive evaluation on our design choices.

	\noindent \textbf{Frame Sampling Strategy in Training}. We compare three frame sampling strategies for obtaining each $\tilde{V}^{(j)}$ in training: uniform sampling, random sampling, and TSN sampling (random sampling from equally divided segments).  
	\vspace{0.5mm}
	\setlength{\columnsep}{9pt}
	\begin{wraptable}{r}[0cm]{0pt}
		\setlength{\tabcolsep}{6pt}
		\renewcommand{\arraystretch}{1.2}
		\hspace{-2.4mm}
		\centering
		\footnotesize
		\begin{tabular}{c|c}
			\specialrule{0em}{0pt}{-10pt}
			Strategy  & AUC (\%) \\
			\shline
			Random~\cite{boris2024surprising} & 83.63 \\
			TSN~\cite{wang2016temporal} & 82.63  \\
			Uniform~\cite{zhang2023video} & \textbf{86.55}    \\
			\specialrule{0em}{-7pt}{0pt}
		\end{tabular}
		\caption{\footnotesize Sampling strategies explored in {\name} training.}
		\label{table:sampling}
		\vspace{-1mm}
	\end{wraptable}
	Table~\ref{table:sampling} shows that uniform sampling performs the best (with batch size $n=$ 2 and $S=$ 8). This is because uniform sampling preserves the temporal structure and maintains consistent motion patterns throughout the long video, making it easier for VLMs to understand the video and update $\mathbf{Q}$.

	\begin{table}[h]
		\setlength{\tabcolsep}{2.2pt}
		\renewcommand{\arraystretch}{1.1}
		\centering
		\footnotesize
		\begin{tabular}{l|c}
			\specialrule{0em}{0pt}{-8pt}
			Question Type  & AUC (\%) \\
			\shline
			No questions & 78.81  \\
			Manually written questions by human & 81.15    \\
			Learned questions w/o iteratively inputting $V_{\rm batch}$ in Eq.~\eqref{eq:optimize}& 78.06 \\\rowcolor{Gray}
			Iteratively learned questions (used in {\name}) & \textbf{86.55} \\
			\specialrule{0em}{-7pt}{0pt}
		\end{tabular}
		\caption{\footnotesize The way we obtain guiding questions affects AUC substantially. }
		\label{table:question}
		\vspace{-2mm}
	\end{table}
	
	\noindent\textbf{How to Obtain Guiding Questions $\mathbf{Q}$ for VLM}. As seen in Table~\ref{table:question}, if the guiding questions are not incorporated into the VLM prompt, the AUC will drop largely to 78.81\%, confirming the need to use simpler and more focused questions to provoke reasoning in the VLMs for VAD. Meanwhile, if we use manually written questions ($\mathbf{Q}_{0}$), the performance is suboptimal with an 81.15\% AUC, which shows the need to use VL to find guiding questions. Lastly, if we only input batched predictions $\hat{Y}_{batch}$ and ground truths $Y_{batch}$ without inputting $V_{\rm batch}$ in the optimizer, the $\mathbf{Q}$ updated in this way will dumb the VLMs and make it have a low AUC. Thus, inputting video frames as Eq.~\eqref{eq:optimize} does is necessary to learn good $\mathbf{Q}$.

	\vspace{0.5mm}

	\begin{wrapfigure}{r}{0.52\linewidth}
		\centering
		\vspace{-3mm}
		\includegraphics[width=1\linewidth]{ 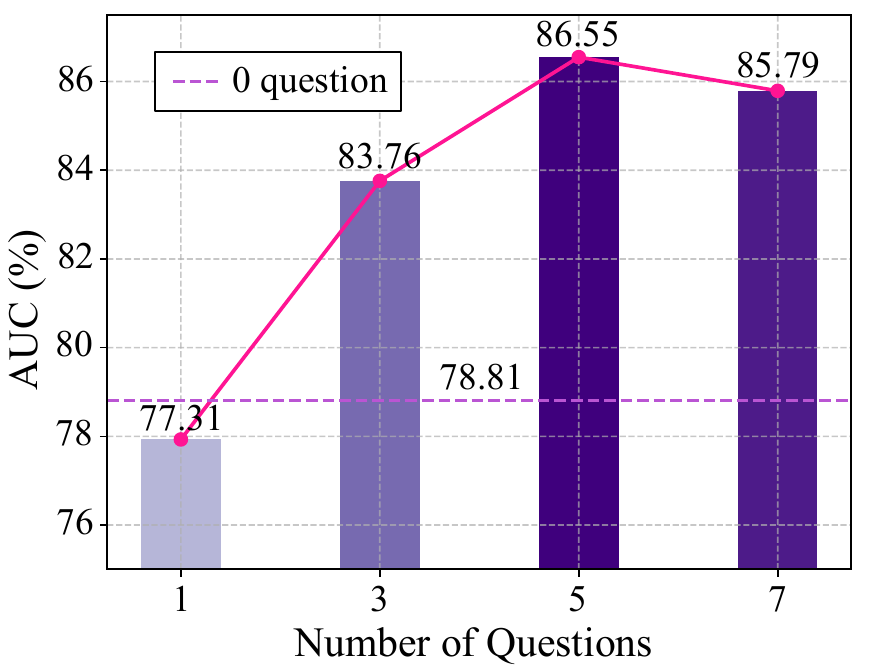}
		\vspace{-7mm}
		\caption{\footnotesize Effect of the number of guiding questions on AUC.}
		\label{table:number}
		\vspace{-3mm}
	\end{wrapfigure}
	
	\noindent\textbf{Number of Questions $m$}.
	As shown in Fig.~\ref{table:number}, when $m$ is set to 1, the reasoning is limited to a single perspective, resulting in a lower AUC. As $m$ increases up to 5, the model captures more comprehensive anomaly patterns, leading to improved AUC. However, increasing $m$ beyond 5 yields no significant gains. Therefore, we set $m$ to 5 by default in {\name}, if not otherwise specified.

	\begin{table}[h]
		\footnotesize
		\setlength{\tabcolsep}{5pt}
		\renewcommand{\arraystretch}{1.1}
		\centering
		\vspace{-2.5mm}
		\begin{tabular}{l|c}
			Operation & AUC (\%) \\
			\shline
			Initial (Step 1) &  76.10\\
			Initial + Retrieval (Step 2)  &  84.53~({\color{darkspringgreen}+8.43})   \\
			Initial + Retrieval + Smoothing (Step 3)  & 85.48~({\color{darkspringgreen}+0.95}) \\\rowcolor{Gray}
			Initial + Retrieval + Smoothing   + Weighting (Step 3) & \textbf{86.55}~({\color{darkspringgreen}+1.07}) \\
		\end{tabular}
		\vspace{-3mm}
		\caption{\footnotesize Ablation study of each step in VERA's inference.}
		\label{table:score}
		\vspace{-2.5mm}
	\end{table}

	\vspace{0.5mm}
	
	\noindent\textbf{Coarse-to-Fine Anomaly Score Computation}. We also validate the anomaly score computation by {\name}. Table~\ref{table:score} shows the AUC is 76.10\% when using the flattened initial score obtained in Step 1, and leveraging retrieved segments in Step 2 significantly boosts the AUC to 84.53\%, highlighting the effectiveness of incorporating ensemble scores based on scene context. Meanwhile, smoothing and weighting in Step 3 further improves the AUC by around 1\% each, verifying the benefit of integrating temporal context.

	\noindent\textbf{Generalizability Test}. We further examine the generalizability of {\name} across different model sizes, VLM architectures, and datasets to address Q3. 
	\vspace{0.5mm}
	\setlength{\columnsep}{8.5pt}
	\begin{wraptable}{r}[0cm]{0pt}
		\centering
		\footnotesize
		\hspace{-2.4mm} 
		\setlength{\tabcolsep}{1.25pt}
		\renewcommand{\arraystretch}{1.1}
		\begin{tabular}{c|cc}
			\specialrule{0em}{0pt}{-13pt}
			\multirow{2}{*}{$f_{\rm VLM}$} & \multicolumn{2}{c}{Source of $\mathbf{Q}$}\\
			& {\scriptsize InternVL2-8B} & {\scriptsize InternVL2-40B}  \\
			\shline
			{\scriptsize InternVL2-8B} &  \textbf{86.55} & 80.43 \\	
			{\scriptsize InternVL2-40B} & 85.24 &  \textbf{86.72} \\
			\specialrule{0em}{-8.5pt}{0pt}
		\end{tabular}
		\caption{\footnotesize  AUC (\%) across model sizes.}
		\label{table:transfer_scale}

		\centering
		\footnotesize
		\setlength{\tabcolsep}{1.25pt}
		\renewcommand{\arraystretch}{1.1}
		\begin{tabular}{c|cc}
			\specialrule{0em}{0pt}{2.5pt}
			\multirow{2}{*}{$f_{\rm VLM}$} &    \multicolumn{2}{c}{Source of $\mathbf{Q}$}\\
			&  {\scriptsize InternVL2-8B} & {\scriptsize Qwen2-VL-7B}  \\
			\shline
			{\scriptsize InternVL2-8B} &  \textbf{86.55} & 81.37 \\
			{\scriptsize Qwen2-VL-7B} & 79.60 &   \textbf{82.64}  \\
			\specialrule{0em}{-9pt}{0pt}
		\end{tabular}
		\caption{\footnotesize AUC (\%) across architectures.}
		\label{table:transfer_model}

		\setlength{\tabcolsep}{1.5pt}
		\renewcommand{\arraystretch}{1.1}
		\centering
		\footnotesize
		\begin{tabular}{c|cc}
			\specialrule{0em}{0pt}{2.5pt}
			\multirow{2}{*}{Dataset} & \multicolumn{2}{c}{Source of $\mathbf{Q}$}\\
			& UCF-Crime & XD-Violence  \\
			\shline
			UCF-Crime &  \textbf{86.55} & 80.42 \\
			XD-Violence & 86.26 &  
			\textbf{88.26} \\
			\specialrule{0em}{-9pt}{0pt}
		\end{tabular}
		\caption{\footnotesize AUC (\%) across datasets.}
		\label{table:transfer_dataset} 
		
		\vspace{-3mm}
	\end{wraptable}
	First, we apply {\name} to InternVL2-40B, a larger model in the InternVL2 family compared to InternVL2-8B. As shown in Table~\ref{table:transfer_scale}, InternVL2-40B achieves effective AUC performance, slightly exceeding that of InternVL2-8B, indicating that VL in {\name} enables models of various scales to identify a $\mathbf{Q}$ suitable for their reasoning capabilities. Additionally, We also evaluate the transferability of $\mathbf{Q}$ across different scales and and observe an interesting phenomenon: the $\mathbf{Q}$ learned by InternVL2-8B remains effective for InternVL2-40B, but not vice versa. This is likely because the $\mathbf{Q}$ learned by the smaller model is readily interpretable by the larger model, whereas the $\mathbf{Q}$ derived from the larger model is more complex in syntactic structure and does not align well with the reasoning framework of the smaller model. Secondly, we select a different VLM, Qwen2-VL-7B~\cite{Qwen2VL}, as the backbone for {\name}. As shown in Table~\ref{table:transfer_model}, while the AUC achieved with Qwen2-VL-7B is lower than that with InternVL2-8B, the VL in {\name} remains effective, allowing it to outperform notable baselines such as LAVAD~\cite{zanella2024harnessing}. However, a notable gap exists when transferring $\mathbf{Q}$ across different model architectures in Table~\ref{table:transfer_model}. Developing a universal $\mathbf{Q}$ that can effectively elicit reasoning capabilities across various VLM structures would be an promising direction for future research. Lastly, we observe that the transferability of $\mathbf{Q}$ depends on the training dataset. From Table~\ref{table:transfer_dataset}, we observe that transferring $\mathbf{Q}$ learned from UCF-Crime to XD-Violence results in a smaller performance drop compared to the reverse case. This suggests the source dataset is crucial to the transferability of $\mathbf{Q}$ across datasets.

	\subsection{Qualitative Results and Case Studies}
	\label{sec:case}
	
	\begin{figure}[t]
		\centering\includegraphics[width=0.95\linewidth]{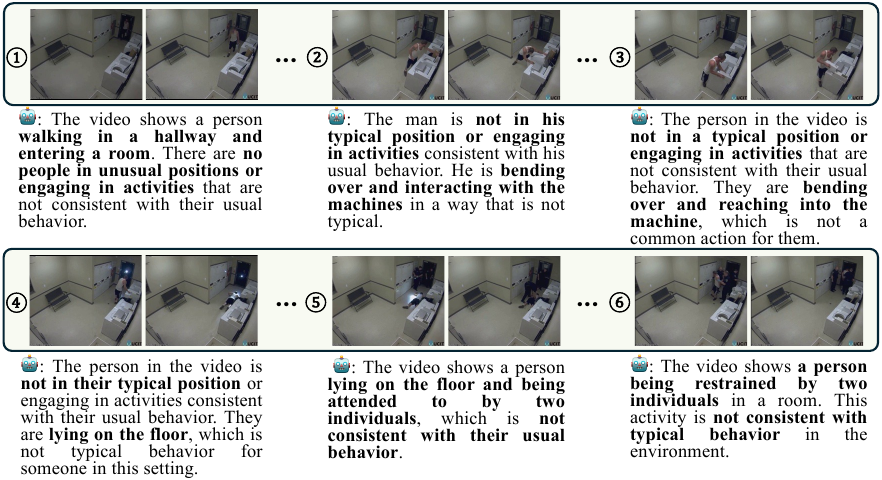}
		\vspace{-2mm}
		\caption{\footnotesize Given $\mathbf{Q}^{*}$ by {\name}, the frozen VLM (InternVL2-8B) will reason and explain the scene based on it. For illustration, we take as an example the video ``Arrest007\_x264" from UCF-Crime and include 6 scenes here. The complete anomaly scores are shown in Fig.~\ref{fig:case}.}
		\label{fig:case_explain}
		\vspace{-2.5mm}
	\end{figure}
\begin{figure}[t]
 
		\centering\includegraphics[width=0.9\linewidth]{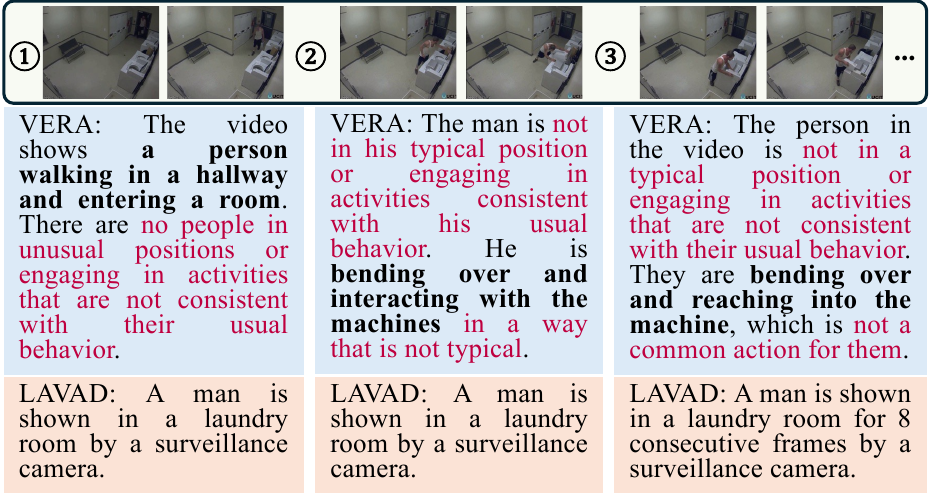}
		\vspace{-3.5mm}
		\caption{\footnotesize Qualitative comparison between VERA and LAVAD.}
		\label{fig:caption_compare}
		\vspace{-3mm}
	\end{figure}

	W.l.o.g., we take one video on UCF-Crime to illustrate the explainability brought by the learned $\mathbf{Q}^{*}$ qualitatively (on UCF-Crime $\mathbf{Q}^{*}$ is ``\emph{1. Are there any people in the video who are not in their typical positions or engaging in activities that are not consistent with their usual behavior? 2. Are there any vehicles in the video that are not in their typical positions or being used in a way that is not consistent with their usual function? 3. Are there any objects in the video that are not in their typical positions or being used in a way that is not consistent with their usual function? 4. Is there any visible damage or unusual movement in the video that indicates an anomaly? 5. Are there any unusual sounds or noises in the video that suggest an anomaly?}''). As shown in Fig.~\ref{fig:case_explain}, the main anomaly in this video is that a man tries to steal money from the washing machines in a laundromat and is arrested after being found by the police. In the selected 6 main video segments, the frozen VLM with {\name}'s learned questions is able to explain the scene by closely following the detailed anomaly characterization of the five learned guiding questions. W.l.o.g., we take the first 3 segments in Fig.~\ref{fig:case_explain} for instance to closely compare the caption quality with LAVAD, a representative baseline. As shown in Fig.~\ref{fig:caption_compare}, VERA's captions include both precise descriptions (\textbf{bold} text) and reasoning (text in  \textcolor{purple}{purple}) about anomalies, while LAVAD's captions contain only plain descriptions. This difference owes to VERA's learned guiding questions, which transform VLM's thinking and phrasing.

\begin{figure}[t]
		\centering
		\vspace{.5mm}
		\includegraphics[width=0.9\linewidth]{ 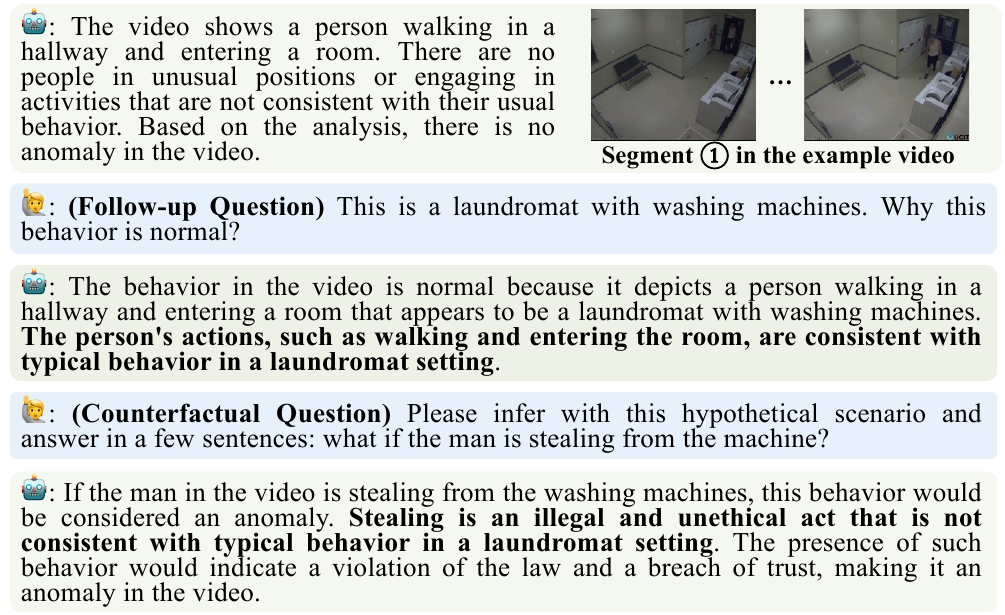}
		\vspace{-2mm}
		\caption{\footnotesize {\name} can take open-ended questions and interact with humans.}
		\label{fig:case_2}
		\vspace{-4mm}
	\end{figure}

	A more interesting advantage of {\name} is that it allows humans to further interact with VLMs because it retains the general question-answering ability of pretrained VLMs. This is because {\name} does not require finetuning of the VLM backbone weights. Although finetuning VLMs with parameter-efficient methods like \cite{hu2022lora,qiu2023controlling,liu2024parameter} is easy and computationally tractable, instruction-tuned models still inevitably lose the flexibility to handle general questions (due to catastrophic forgetting), as they are trained to respond to certain queries with fixed answer styles. In contrast, as shown in Fig.~\ref{fig:case_2}, the learned $\mathbf{Q}^{*}$ can steer reasoning in a frozen VLM while allowing it to flexibly answer open-ended (like follow-up or counterfactual) questions, which is an important ability lost in IT-based models.
	
\begin{wrapfigure}{r}{0.54\linewidth}
		\centering
		\vspace{-3.5mm}
		\includegraphics[width=.97\linewidth]{ 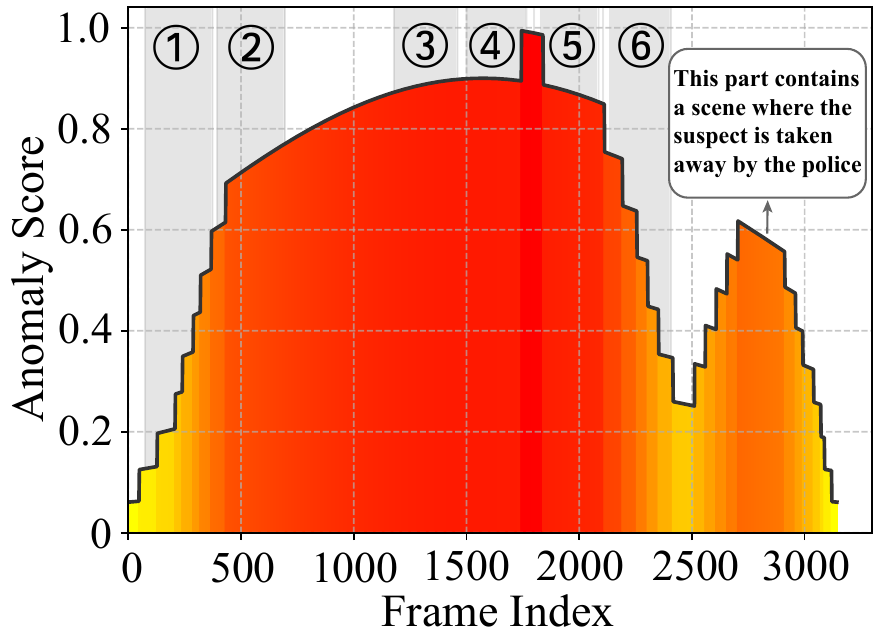}
		\vspace{-3mm}
		\caption{\footnotesize Anomaly scores generated by {\name} (with InternVL2-8B)  in ``Arrest007\_x264" from UCF-Crime.}
		\label{fig:case}
		\vspace{-3mm}
	\end{wrapfigure}

	Moreover, as shown in Fig.~\ref{fig:case}, owing to the proposed 
	\vspace{0.4mm}
	coarse-to-fine anomaly scoring, the anomaly score dynamics from VERA well represent the actual real-time anomaly level in this video and gradually increases to nearly 1 when the man is being arrested. This result verifies that {\name} allows VLMs to effectively identify anomalies with a holistic model, reducing the manpower and computational overhead for VAD.

	\vspace{-.5mm}
	\section{Concluding Remarks}
	\vspace{-.25mm}
	We propose a novel pipeline, {\name}, which can effectively elicit the reasoning ability from VLMs to perform explainable VAD without additional computation overhead. This is done through an effective and novel application of verbalized machine learning~\cite{xiao2024verbalized} to VLM. In training, {\name} obtains the guiding questions detailing anomaly patterns through the verbal interaction between the learner and the optimizer agents. In inference, {\name} uses them to enhance VLMs for identifying anomalies and compute frame-level anomaly scores in a coarse-to-fine process. Experimental results validate the effectiveness of the {\name} framework in achieving state-of-the-art explainable VAD performance.

	{
		\small
		\bibliographystyle{ieeenat_fullname}
		\bibliography{main}

\begin{thebibliography}{58}
\providecommand{\natexlab}[1]{#1}
\providecommand{\url}[1]{\texttt{#1}}
\expandafter\ifx\csname urlstyle\endcsname\relax
  \providecommand{\doi}[1]{doi: #1}\else
  \providecommand{\doi}{doi: \begingroup \urlstyle{rm}\Url}\fi

\bibitem[Achiam et~al.(2023)Achiam, Adler, Agarwal, Ahmad, Akkaya, Aleman,
  Almeida, Altenschmidt, Altman, Anadkat, et~al.]{achiam2023gpt}
Josh Achiam, Steven Adler, Sandhini Agarwal, Lama Ahmad, Ilge Akkaya,
  Florencia~Leoni Aleman, Diogo Almeida, Janko Altenschmidt, Sam Altman,
  Shyamal Anadkat, et~al.
\newblock Gpt-4 technical report.
\newblock \emph{arXiv preprint arXiv:2303.08774}, 2023.

\bibitem[Bogdoll et~al.(2022)Bogdoll, Nitsche, and
  Z{\"o}llner]{bogdoll2022anomaly}
Daniel Bogdoll, Maximilian Nitsche, and J~Marius Z{\"o}llner.
\newblock Anomaly detection in autonomous driving: A survey.
\newblock In \emph{CVPR Workshops}, 2022.

\bibitem[Boris et~al.(2024)Boris, Anil, Anna, and Marcus]{boris2024surprising}
Meinardus Boris, Batra Anil, Rohrbach Anna, and Rohrbach Marcus.
\newblock The surprising effectiveness of multimodal large language models for
  video moment retrieval.
\newblock \emph{arXiv preprint arXiv:2406.18113}, 2024.

\bibitem[Brown et~al.(2020)Brown, Mann, Ryder, Subbiah, Kaplan, Dhariwal,
  Neelakantan, Shyam, Sastry, Askell, Agarwal, Herbert{-}Voss, Krueger,
  Henighan, Child, Ramesh, Ziegler, Wu, Winter, Hesse, Chen, Sigler, Litwin,
  Gray, Chess, Clark, Berner, McCandlish, Radford, Sutskever, and
  Amodei]{DBLP:conf/nips/BrownMRSKDNSSAA20}
Tom~B. Brown, Benjamin Mann, Nick Ryder, Melanie Subbiah, Jared Kaplan,
  Prafulla Dhariwal, Arvind Neelakantan, Pranav Shyam, Girish Sastry, Amanda
  Askell, Sandhini Agarwal, Ariel Herbert{-}Voss, Gretchen Krueger, Tom
  Henighan, Rewon Child, Aditya Ramesh, Daniel~M. Ziegler, Jeffrey Wu, Clemens
  Winter, Christopher Hesse, Mark Chen, Eric Sigler, Mateusz Litwin, Scott
  Gray, Benjamin Chess, Jack Clark, Christopher Berner, Sam McCandlish, Alec
  Radford, Ilya Sutskever, and Dario Amodei.
\newblock Language models are few-shot learners.
\newblock In \emph{NeurIPS}, 2020.

\bibitem[Chen et~al.(2024{\natexlab{a}})Chen, Li, Su, Zha, and
  Huang]{chen2024prompt}
Junxi Chen, Liang Li, Li Su, Zheng-jun Zha, and Qingming Huang.
\newblock Prompt-enhanced multiple instance learning for weakly supervised
  video anomaly detection.
\newblock In \emph{CVPR}, 2024{\natexlab{a}}.

\bibitem[Chen et~al.(2023)Chen, Liu, Zhang, Fok, Qi, and Wu]{chen2023mgfn}
Yingxian Chen, Zhengzhe Liu, Baoheng Zhang, Wilton Fok, Xiaojuan Qi, and
  Yik-Chung Wu.
\newblock Mgfn: Magnitude-contrastive glance-and-focus network for
  weakly-supervised video anomaly detection.
\newblock In \emph{AAAI}, 2023.

\bibitem[Chen et~al.(2024{\natexlab{b}})Chen, Wu, Wang, Su, Chen, Xing, Zhong,
  Zhang, Zhu, Lu, et~al.]{chen2024internvl}
Zhe Chen, Jiannan Wu, Wenhai Wang, Weijie Su, Guo Chen, Sen Xing, Muyan Zhong,
  Qinglong Zhang, Xizhou Zhu, Lewei Lu, et~al.
\newblock Internvl: Scaling up vision foundation models and aligning for
  generic visual-linguistic tasks.
\newblock In \emph{CVPR}, 2024{\natexlab{b}}.

\bibitem[Dosovitskiy et~al.(2021)Dosovitskiy, Beyer, Kolesnikov, Weissenborn,
  Zhai, Unterthiner, Dehghani, Minderer, Heigold, Gelly, Uszkoreit, and
  Houlsby]{DBLP:conf/iclr/DosovitskiyB0WZ21}
Alexey Dosovitskiy, Lucas Beyer, Alexander Kolesnikov, Dirk Weissenborn,
  Xiaohua Zhai, Thomas Unterthiner, Mostafa Dehghani, Matthias Minderer, Georg
  Heigold, Sylvain Gelly, Jakob Uszkoreit, and Neil Houlsby.
\newblock An image is worth 16x16 words: Transformers for image recognition at
  scale.
\newblock In \emph{ICLR}, 2021.

\bibitem[Feng et~al.(2021)Feng, Hong, and Zheng]{feng2021mist}
Jia-Chang Feng, Fa-Ting Hong, and Wei-Shi Zheng.
\newblock Mist: Multiple instance self-training framework for video anomaly
  detection.
\newblock In \emph{CVPR}, 2021.

\bibitem[Girdhar et~al.(2023)Girdhar, El-Nouby, Liu, Singh, Alwala, Joulin, and
  Misra]{girdhar2023imagebind}
Rohit Girdhar, Alaaeldin El-Nouby, Zhuang Liu, Mannat Singh, Kalyan~Vasudev
  Alwala, Armand Joulin, and Ishan Misra.
\newblock Imagebind: One embedding space to bind them all.
\newblock In \emph{CVPR}, 2023.

\bibitem[Gonzalez(2009)]{gonzalez2009digital}
Rafael~C Gonzalez.
\newblock \emph{Digital image processing}.
\newblock Pearson education india, 2009.

\bibitem[Guo et~al.(2024)Guo, De~Mello, Yin, Byeon, Cheung, Yu, Luo, and
  Liu]{guo2024regiongpt}
Qiushan Guo, Shalini De~Mello, Hongxu Yin, Wonmin Byeon, Ka~Chun Cheung, Yizhou
  Yu, Ping Luo, and Sifei Liu.
\newblock Regiongpt: Towards region understanding vision language model.
\newblock In \emph{CVPR}, 2024.

\bibitem[Hasan et~al.(2016)Hasan, Choi, Neumann, Roy-Chowdhury, and
  Davis]{hasan2016learning}
Mahmudul Hasan, Jonghyun Choi, Jan Neumann, Amit~K Roy-Chowdhury, and Larry~S
  Davis.
\newblock Learning temporal regularity in video sequences.
\newblock In \emph{CVPR}, 2016.

\bibitem[Hu et~al.(2021)Hu, Wallis, Allen-Zhu, Li, Wang, Wang, Chen,
  et~al.]{hu2022lora}
Edward~J Hu, Phillip Wallis, Zeyuan Allen-Zhu, Yuanzhi Li, Shean Wang, Lu Wang,
  Weizhu Chen, et~al.
\newblock Lora: Low-rank adaptation of large language models.
\newblock In \emph{ICLR}, 2021.

\bibitem[Joo et~al.(2023)Joo, Vo, Yamazaki, and Le]{joo2023clip}
Hyekang~Kevin Joo, Khoa Vo, Kashu Yamazaki, and Ngan Le.
\newblock Clip-tsa: Clip-assisted temporal self-attention for weakly-supervised
  video anomaly detection.
\newblock In \emph{ICIP}, 2023.

\bibitem[Kingma(2014)]{kingma2014adam}
Diederik~P Kingma.
\newblock Adam: A method for stochastic optimization.
\newblock \emph{arXiv preprint arXiv:1412.6980}, 2014.

\bibitem[Li et~al.(2022{\natexlab{a}})Li, Cai, Zeng, and Zhao]{li2022scale}
Guoqiu Li, Guanxiong Cai, Xingyu Zeng, and Rui Zhao.
\newblock Scale-aware spatio-temporal relation learning for video anomaly
  detection.
\newblock In \emph{ECCV}, 2022{\natexlab{a}}.

\bibitem[Li et~al.(2023)Li, Li, Savarese, and Hoi]{li2023blip}
Junnan Li, Dongxu Li, Silvio Savarese, and Steven Hoi.
\newblock Blip-2: Bootstrapping language-image pre-training with frozen image
  encoders and large language models.
\newblock In \emph{ICML}, 2023.

\bibitem[Li et~al.(2022{\natexlab{b}})Li, Liu, and Jiao]{li2022self}
Shuo Li, Fang Liu, and Licheng Jiao.
\newblock Self-training multi-sequence learning with transformer for weakly
  supervised video anomaly detection.
\newblock In \emph{AAAI}, 2022{\natexlab{b}}.

\bibitem[Liu et~al.(2024{\natexlab{a}})Liu, Li, Li, and Lee]{liu2024improved}
Haotian Liu, Chunyuan Li, Yuheng Li, and Yong~Jae Lee.
\newblock Improved baselines with visual instruction tuning.
\newblock In \emph{CVPR}, 2024{\natexlab{a}}.

\bibitem[Liu et~al.(2024{\natexlab{b}})Liu, Li, Wu, and Lee]{liu2024visual}
Haotian Liu, Chunyuan Li, Qingyang Wu, and Yong~Jae Lee.
\newblock Visual instruction tuning.
\newblock In \emph{NeurIPS}, 2024{\natexlab{b}}.

\bibitem[Liu et~al.(2017)Liu, Zhang, Li, Yu, Dai, Zhao, and Song]{liu2017deep}
Weiyang Liu, Yan-Ming Zhang, Xingguo Li, Zhiding Yu, Bo Dai, Tuo Zhao, and Le
  Song.
\newblock Deep hyperspherical learning.
\newblock In \emph{NeurIPS}, 2017.

\bibitem[Liu et~al.(2018)Liu, Luo, Lian, and Gao]{liu2018future}
Wen Liu, Weixin Luo, Dongze Lian, and Shenghua Gao.
\newblock Future frame prediction for anomaly detection--a new baseline.
\newblock In \emph{CVPR}, 2018.

\bibitem[Liu et~al.(2024{\natexlab{c}})Liu, Qiu, Feng, Xiu, Xue, Yu, Feng, Liu,
  Heo, Peng, et~al.]{liu2024parameter}
Weiyang Liu, Zeju Qiu, Yao Feng, Yuliang Xiu, Yuxuan Xue, Longhui Yu, Haiwen
  Feng, Zhen Liu, Juyeon Heo, Songyou Peng, et~al.
\newblock Parameter-efficient orthogonal finetuning via butterfly
  factorization.
\newblock In \emph{ICLR}, 2024{\natexlab{c}}.

\bibitem[Lu et~al.(2013)Lu, Shi, and Jia]{lu2013abnormal}
Cewu Lu, Jianping Shi, and Jiaya Jia.
\newblock Abnormal event detection at 150 fps in matlab.
\newblock In \emph{ICCV}, 2013.

\bibitem[Lv and Sun(2024)]{lv2024video}
Hui Lv and Qianru Sun.
\newblock Video anomaly detection and explanation via large language models.
\newblock \emph{arXiv preprint arXiv:2401.05702}, 2024.

\bibitem[Lv et~al.(2023)Lv, Yue, Sun, Luo, Cui, and Zhang]{lv2023unbiased}
Hui Lv, Zhongqi Yue, Qianru Sun, Bin Luo, Zhen Cui, and Hanwang Zhang.
\newblock Unbiased multiple instance learning for weakly supervised video
  anomaly detection.
\newblock In \emph{CVPR}, 2023.

\bibitem[Pratt et~al.(2023)Pratt, Covert, Liu, and Farhadi]{pratt2023does}
Sarah Pratt, Ian Covert, Rosanne Liu, and Ali Farhadi.
\newblock What does a platypus look like? generating customized prompts for
  zero-shot image classification.
\newblock In \emph{ICCV}, 2023.

\bibitem[Qiu et~al.(2023)Qiu, Liu, Feng, Xue, Feng, Liu, Zhang, Weller, and
  Sch{\"o}lkopf]{qiu2023controlling}
Zeju Qiu, Weiyang Liu, Haiwen Feng, Yuxuan Xue, Yao Feng, Zhen Liu, Dan Zhang,
  Adrian Weller, and Bernhard Sch{\"o}lkopf.
\newblock Controlling text-to-image diffusion by orthogonal finetuning.
\newblock In \emph{NeurIPS}, 2023.

\bibitem[Radford et~al.(2021)Radford, Kim, Hallacy, Ramesh, Goh, Agarwal,
  Sastry, Askell, Mishkin, Clark, et~al.]{radford2021learning}
Alec Radford, Jong~Wook Kim, Chris Hallacy, Aditya Ramesh, Gabriel Goh,
  Sandhini Agarwal, Girish Sastry, Amanda Askell, Pamela Mishkin, Jack Clark,
  et~al.
\newblock Learning transferable visual models from natural language
  supervision.
\newblock In \emph{ICML}, 2021.

\bibitem[Roth et~al.(2022)Roth, Pemula, Zepeda, Sch{\"o}lkopf, Brox, and
  Gehler]{roth2022towards}
Karsten Roth, Latha Pemula, Joaquin Zepeda, Bernhard Sch{\"o}lkopf, Thomas
  Brox, and Peter Gehler.
\newblock Towards total recall in industrial anomaly detection.
\newblock In \emph{CVPR}, 2022.

\bibitem[Sultani et~al.(2018)Sultani, Chen, and Shah]{sultani2018real}
Waqas Sultani, Chen Chen, and Mubarak Shah.
\newblock Real-world anomaly detection in surveillance videos.
\newblock In \emph{CVPR}, 2018.

\bibitem[Tang et~al.(2024)Tang, Lu, Wu, Xu, Ma, Fang, Guo, Lu, Chen, and
  Chen]{tang2024hawk}
Jiaqi Tang, Hao Lu, Ruizheng Wu, Xiaogang Xu, Ke Ma, Cheng Fang, Bin Guo,
  Jiangbo Lu, Qifeng Chen, and Ying-Cong Chen.
\newblock Hawk: Learning to understand open-world video anomalies.
\newblock \emph{arXiv preprint arXiv:2405.16886}, 2024.

\bibitem[Thakare et~al.(2023{\natexlab{a}})Thakare, Dogra, Choi, Kim, and
  Kim]{thakare2023rareanom}
Kamalakar~Vijay Thakare, Debi~Prosad Dogra, Heeseung Choi, Haksub Kim, and
  Ig-Jae Kim.
\newblock Rareanom: A benchmark video dataset for rare type anomalies.
\newblock \emph{Pattern Recognition}, 140:\penalty0 109567, 2023{\natexlab{a}}.

\bibitem[Thakare et~al.(2023{\natexlab{b}})Thakare, Raghuwanshi, Dogra, Choi,
  and Kim]{thakare2023dyannet}
Kamalakar~Vijay Thakare, Yash Raghuwanshi, Debi~Prosad Dogra, Heeseung Choi,
  and Ig-Jae Kim.
\newblock Dyannet: A scene dynamicity guided self-trained video anomaly
  detection network.
\newblock In \emph{WACV}, 2023{\natexlab{b}}.

\bibitem[Tian et~al.(2021)Tian, Pang, Chen, Singh, Verjans, and
  Carneiro]{tian2021weakly}
Yu Tian, Guansong Pang, Yuanhong Chen, Rajvinder Singh, Johan~W Verjans, and
  Gustavo Carneiro.
\newblock Weakly-supervised video anomaly detection with robust temporal
  feature magnitude learning.
\newblock In \emph{ICCV}, 2021.

\bibitem[Tur et~al.(2023)Tur, Dall’Asen, Beyan, and
  Ricci]{tur2023unsupervised}
Anil~Osman Tur, Nicola Dall’Asen, Cigdem Beyan, and Elisa Ricci.
\newblock Unsupervised video anomaly detection with diffusion models
  conditioned on compact motion representations.
\newblock In \emph{International Conference on Image Analysis and Processing},
  2023.

\bibitem[Wang and Cherian(2019)]{wang2019gods}
Jue Wang and Anoop Cherian.
\newblock Gods: Generalized one-class discriminative subspaces for anomaly
  detection.
\newblock In \emph{ICCV}, 2019.

\bibitem[Wang et~al.(2016)Wang, Xiong, Wang, Qiao, Lin, Tang, and
  Van~Gool]{wang2016temporal}
Limin Wang, Yuanjun Xiong, Zhe Wang, Yu Qiao, Dahua Lin, Xiaoou Tang, and Luc
  Van~Gool.
\newblock Temporal segment networks: Towards good practices for deep action
  recognition.
\newblock In \emph{ECCV}, 2016.

\bibitem[Wang et~al.(2024)Wang, Bai, Tan, Wang, Fan, Bai, Chen, Liu, Wang, Ge,
  Fan, Dang, Du, Ren, Men, Liu, Zhou, Zhou, and Lin]{Qwen2VL}
Peng Wang, Shuai Bai, Sinan Tan, Shijie Wang, Zhihao Fan, Jinze Bai, Keqin
  Chen, Xuejing Liu, Jialin Wang, Wenbin Ge, Yang Fan, Kai Dang, Mengfei Du,
  Xuancheng Ren, Rui Men, Dayiheng Liu, Chang Zhou, Jingren Zhou, and Junyang
  Lin.
\newblock Qwen2-vl: Enhancing vision-language model's perception of the world
  at any resolution.
\newblock \emph{arXiv preprint arXiv:2409.12191}, 2024.

\bibitem[Wu et~al.(2022)Wu, Hsieh, Chen, Fuh, and Liu]{wu2022self}
Jhih-Ciang Wu, He-Yen Hsieh, Ding-Jie Chen, Chiou-Shann Fuh, and Tyng-Luh Liu.
\newblock Self-supervised sparse representation for video anomaly detection.
\newblock In \emph{ECCV}, 2022.

\bibitem[Wu et~al.(2020)Wu, Liu, Shi, Sun, Shao, Wu, and Yang]{wu2020not}
Peng Wu, Jing Liu, Yujia Shi, Yujia Sun, Fangtao Shao, Zhaoyang Wu, and Zhiwei
  Yang.
\newblock Not only look, but also listen: Learning multimodal violence
  detection under weak supervision.
\newblock In \emph{ECCV}, 2020.

\bibitem[Wu et~al.(2024{\natexlab{a}})Wu, Zhou, Pang, Sun, Liu, Wang, and
  Zhang]{wu2024open}
Peng Wu, Xuerong Zhou, Guansong Pang, Yujia Sun, Jing Liu, Peng Wang, and
  Yanning Zhang.
\newblock Open-vocabulary video anomaly detection.
\newblock In \emph{CVPR}, pages 18297--18307, 2024{\natexlab{a}}.

\bibitem[Wu et~al.(2024{\natexlab{b}})Wu, Zhou, Pang, Zhou, Yan, Wang, and
  Zhang]{wu2024vadclip}
Peng Wu, Xuerong Zhou, Guansong Pang, Lingru Zhou, Qingsen Yan, Peng Wang, and
  Yanning Zhang.
\newblock Vadclip: Adapting vision-language models for weakly supervised video
  anomaly detection.
\newblock In \emph{Proceedings of the AAAI Conference on Artificial
  Intelligence}, pages 6074--6082, 2024{\natexlab{b}}.

\bibitem[Xiao et~al.(2024)Xiao, Bamler, Sch{\"o}lkopf, and
  Liu]{xiao2024verbalized}
Tim~Z Xiao, Robert Bamler, Bernhard Sch{\"o}lkopf, and Weiyang Liu.
\newblock Verbalized machine learning: Revisiting machine learning with
  language models.
\newblock \emph{arXiv preprint arXiv:2406.04344}, 2024.

\bibitem[Yang et~al.(2024{\natexlab{a}})Yang, Lee, Dariush, Cao, and
  Lo]{yang2024follow}
Yuchen Yang, Kwonjoon Lee, Behzad Dariush, Yinzhi Cao, and Shao-Yuan Lo.
\newblock Follow the rules: reasoning for video anomaly detection with large
  language models.
\newblock \emph{arXiv preprint arXiv:2407.10299}, 2024{\natexlab{a}}.

\bibitem[Yang et~al.(2024{\natexlab{b}})Yang, Liu, and Wu]{yang2024text}
Zhiwei Yang, Jing Liu, and Peng Wu.
\newblock Text prompt with normality guidance for weakly supervised video
  anomaly detection.
\newblock In \emph{CVPR}, 2024{\natexlab{b}}.

\bibitem[Ye et~al.(2019)Ye, Peng, Gan, Wu, and Qiao]{ye2019anopcn}
Muchao Ye, Xiaojiang Peng, Weihao Gan, Wei Wu, and Yu Qiao.
\newblock Anopcn: Video anomaly detection via deep predictive coding network.
\newblock In \emph{ACM international conference on multimedia}, 2019.

\bibitem[Yuksekgonul et~al.(2024)Yuksekgonul, Bianchi, Boen, Liu, Huang,
  Guestrin, and Zou]{yuksekgonul2024textgrad}
Mert Yuksekgonul, Federico Bianchi, Joseph Boen, Sheng Liu, Zhi Huang, Carlos
  Guestrin, and James Zou.
\newblock Textgrad: Automatic" differentiation" via text.
\newblock \emph{arXiv preprint arXiv:2406.07496}, 2024.

\bibitem[Zaheer et~al.(2020)Zaheer, Mahmood, Astrid, and Lee]{zaheer2020claws}
Muhammad~Zaigham Zaheer, Arif Mahmood, Marcella Astrid, and Seung-Ik Lee.
\newblock Claws: Clustering assisted weakly supervised learning with normalcy
  suppression for anomalous event detection.
\newblock In \emph{ECCV}, 2020.

\bibitem[Zaheer et~al.(2022)Zaheer, Mahmood, Khan, Segu, Yu, and
  Lee]{zaheer2022generative}
M~Zaigham Zaheer, Arif Mahmood, M~Haris Khan, Mattia Segu, Fisher Yu, and
  Seung-Ik Lee.
\newblock Generative cooperative learning for unsupervised video anomaly
  detection.
\newblock In \emph{CVPR}, 2022.

\bibitem[Zanella et~al.(2024)Zanella, Menapace, Mancini, Wang, and
  Ricci]{zanella2024harnessing}
Luca Zanella, Willi Menapace, Massimiliano Mancini, Yiming Wang, and Elisa
  Ricci.
\newblock Harnessing large language models for training-free video anomaly
  detection.
\newblock In \emph{CVPR}, 2024.

\bibitem[Zhang et~al.(2023{\natexlab{a}})Zhang, Li, Qi, Wang, Qing, Huang, and
  Yang]{zhang2023exploiting}
Chen Zhang, Guorong Li, Yuankai Qi, Shuhui Wang, Laiyun Qing, Qingming Huang,
  and Ming-Hsuan Yang.
\newblock Exploiting completeness and uncertainty of pseudo labels for weakly
  supervised video anomaly detection.
\newblock In \emph{CVPR}, 2023{\natexlab{a}}.

\bibitem[Zhang et~al.(2023{\natexlab{b}})Zhang, Li, and Bing]{zhang2023video}
Hang Zhang, Xin Li, and Lidong Bing.
\newblock Video-llama: An instruction-tuned audio-visual language model for
  video understanding.
\newblock In \emph{EMNLP}, 2023{\natexlab{b}}.

\bibitem[Zhang et~al.(2024{\natexlab{a}})Zhang, Xu, Wang, Zuo, Han, Huang, Gao,
  Wang, and Sang]{zhang2024holmes}
Huaxin Zhang, Xiaohao Xu, Xiang Wang, Jialong Zuo, Chuchu Han, Xiaonan Huang,
  Changxin Gao, Yuehuan Wang, and Nong Sang.
\newblock Holmes-vad: Towards unbiased and explainable video anomaly detection
  via multi-modal llm.
\newblock \emph{arXiv preprint arXiv:2406.12235}, 2024{\natexlab{a}}.

\bibitem[Zhang et~al.(2024{\natexlab{b}})Zhang, Wang, Qi, Sun, Zhuang, Ren, Ma,
  and Liao]{zhang2024multi}
Menghao Zhang, Jingyu Wang, Qi Qi, Haifeng Sun, Zirui Zhuang, Pengfei Ren,
  Ruilong Ma, and Jianxin Liao.
\newblock Multi-scale video anomaly detection by multi-grained spatio-temporal
  representation learning.
\newblock In \emph{CVPR}, 2024{\natexlab{b}}.

\bibitem[Zhong et~al.(2019)Zhong, Li, Kong, Liu, Li, and Li]{zhong2019graph}
Jia-Xing Zhong, Nannan Li, Weijie Kong, Shan Liu, Thomas~H Li, and Ge Li.
\newblock Graph convolutional label noise cleaner: Train a plug-and-play action
  classifier for anomaly detection.
\newblock In \emph{CVPR}, 2019.

\bibitem[Zhu et~al.(2024)Zhu, Chen, Shen, Li, and Elhoseiny]{zhuminigpt}
Deyao Zhu, Jun Chen, Xiaoqian Shen, Xiang Li, and Mohamed Elhoseiny.
\newblock Minigpt-4: Enhancing vision-language understanding with advanced
  large language models.
\newblock In \emph{ICLR}, 2024.

\end{thebibliography}
	}
	
	\clearpage
	
	\newpage
	\onecolumn
	\addcontentsline{toc}{section}{Appendix} 
	\renewcommand \thepart{} 
	\renewcommand \partname{}
	\part{\Large{\centerline{Appendix}}}
	\parttoc

	\newpage
	\onecolumn
	\appendix
	
	In this supplementary material, we first include more details on training in {\name} (Sec.~\ref{sec:train2}) and additional experimental results (Sec.~\ref{sec:additional_exp}). To specify:
	\begin{itemize}
		\item In Sec.~\ref{sec:train2}, we provide the pseudocodes and details on the initialization, the learner prompt template, and the optimizer prompt template for the training process in Sec.~\ref{sec:pseudocodes}. After that, we discuss the optimization process of the learned questions by the optimizer in Sec.~\ref{sec:optimizer}.
		\item In Sec.~\ref{sec:additional_exp}, we first include comparison results with the state-of-the-art methods on XD-Violence measured by AP in Sec.~\ref{sec:ap}. We also discuss other good properties of {\name}, including the good generalizability of the learned questions for different scenarios and the insensitivity of {\name} regarding hyperparameters in Sec.~\ref{sec:gen_vs_specific} and Sec.~\ref{sec:hyperparams}, respectively. Finally, we include additional case studies with normal and abnormal videos in Sec.~\ref{sec:additional}.
		
	\end{itemize}
	
	We also include a further discussion on the limitations of VERA for future research exploration in Sec.~\ref{sec:limitation}.

	\section{Training in VERA}
	\label{sec:train2}
	\subsection{Algorithm}
	\label{sec:pseudocodes}
	We show the complete iterative training process of {\name} in pseudocodes in Algorithm~\ref{Alg: pseudocode}. It is an iterative process of using the learner to output binary prediction for each sample in a mini-batch and asking the optimizer to update the guiding questions after collecting the batched data. Meanwhile, we have a small validation set (10\% samples randomly drawn from the original training set) for deciding the $\mathbf{Q}^{*}$ used for testing. We want to further detail on certain elements in Algorithm~\ref{Alg: pseudocode} as follows.

	\begin{algorithm}[h]
		
		\small
		\SetAlgoLined
		
		\textbf{Inputs}: Training data pairs $D_{\rm train}=\{(\tilde{V}^{(j)}, Y^{(j)})\}_{j=1}^{N}$, iteration number  $P$, initial guiding questions $\mathbf{Q}_{0}$, learner $f_{\rm learner}$, optimizer $f_{\rm opt}$, learner prompt template $\theta$, optimizer prompt template $\psi$, validation set $D_{\rm val}=\{(\tilde{V}_{\rm val}^{(j)}, Y_{\rm val}^{(j)})\}_{j=1}^{\eta}$, period for validation $\mu$, batch size $n$.
		
		\textbf{Output}: Optimal guiding questions $\mathbf{Q}^{*}$.

		Set iteration counter $t\leftarrow1$;
		
		Set $\mathbf{Q}^{*}\leftarrow \mathbf{Q}_{0}$, test $\mathbf{Q}_{0}$ on validation set $D_{\rm val}$ and compute its validation accuracy as ${\rm Acc}^{*}$;
		
		\While{t $\le$ P }{
			
			\textcolor{blue}{{\# \ \emph{Conduct the learning task with a mini-batch by the learner}}}
			
			Randomly sample a batch without repetition from $D_{\rm train}$ with a visual input batch $V_{\rm batch} = [\tilde{V}_{\rm batch}^{(1)},\cdots,\tilde{V}_{\rm batch}^{(n)}]$ and ground truths $Y_{\rm batch}=[Y_{\rm batch}^{(1)},\cdots,Y_{\rm batch}^{(n)}]$;

			\For{$1 \le j \le n$}{
				Obtain a prediction $\hat{Y}_{\rm batch}^{(j)}$ for $\tilde{V}_{\rm batch}^{(j)}$ from $f_{\rm learner}$ with prompt $(\theta, \mathbf{Q}_{t})$ by Eq.~\eqref{eq:predict} as 
				$\hat{Y}_{\rm batch}^{(j)}= f_{\rm learner}^{(t)}(\tilde{V}_{\rm batch}^{(j)})$;
			}
			
			\textcolor{blue}{{\# \ \emph{Update the guiding questions with the batched data by the optimizer}}}
			
			Input the batched prediction $\hat{Y}_{\rm batch}=[\hat{Y}_{\rm batch}^{(1)},\cdots,\hat{Y}_{\rm batch}^{(n)}]$ with $V_{\rm batch}$  and $Y_{\rm batch}$ into the optimizer for obtaining a new set of guiding questions by Eq.~\eqref{eq:optimize} as:
			$\mathbf{Q}_{t+1} = f_{\rm opt}^{(t)}(V_{\rm batch}, \hat{Y}_{\rm batch}, Y_{\rm batch})$;

			\textcolor{blue}{{\# \ \emph{Compute the validation accuracy with the learned guiding questions periodically}}}
			
			$t\leftarrow t+1$;
			
			\If{$t \ {\rm mod} \ \mu = 0$}{
				Test $\mathbf{Q}_{t}$ on the validation set  $D_{\rm val}$ and compute the validation accuracy ${\rm Acc}_{t}$;
				
				\If{${\rm Acc}_{t}>{\rm Acc}^{*}$}{
					
					Update $\mathbf{Q}^{*}\leftarrow\mathbf{Q}_t$;
					
					Update ${\rm ACC}^{*}\leftarrow{\rm ACC}_{t}$;
				}

			}

		}
		
		\textbf{Return} $\mathbf{Q}^{*}$;
		
		\caption{Optimizing Guiding Questions in VAD by {\name} during Training}
		\label{Alg: pseudocode}
	\end{algorithm}

	\noindent \textbf{Initial $\mathbf{Q}_{0}$}. The initial guiding questions $\mathbf{Q}_{0}$ are ``\emph{1. Is there any suspicious person or object that looks unusual in this scene? 2. Is there any behavior that looks unusual in this scene?}''. These two questions are manually written and inspired by previous VAD methods, which assume anomaly as something or somebody with unusual appearance or motions~\cite{wu2024open,hasan2016learning}. This set of questions is also the ``\textbf{manually written questions by human}'' in Table~\ref{table:question}, which is suboptimal in guiding frozen VLMs to detect anomalies. The key idea of training is to use VL to iteratively update $\mathbf{Q}$ given a suboptimal $\mathbf{Q}_0$.

	\noindent \textbf{Learner Prompt Template $\theta$}.
	We detail the design of $\theta$ as follows. As shown in Fig.~\ref{fig:qvml}, the learner prompt template $\theta$ includes four sections, \ie, Model Description, Prompt Questions, Input, and Output Formatting. To specify:
	\begin{itemize}
		\item Model Description: This section introduces the learning task, providing the learner with the necessary background knowledge to understand the objective. It clarifies what the learner is expected to predict based on the given visual input data.
		\item Prompt Questions: This section presents a general prompt to guide the learner's reasoning process. Specific prompts, denoted as $\mathbf{Q}_{t}$, will be inserted here to facilitate reasoning within a frozen VLM.
		\item Input: This section simply stores the visual tokens. When the VLM reads this, it will correlate the read text with the visual inputs.
		\item Output Formatting: The last section in $\theta$ mainly provides information on output formats to ensure that VLMs think through the given questions $\mathbf{Q}_t$ and output a prediction in a format easy for post-processing in computers. 
	\end{itemize}
	
	\noindent \textbf{Optimizer Prompt Template $\psi$}. As shown in Fig.~\ref{fig:qvml}, the optimizer prompt template includes seven sections, \ie, Instruction, Inputs, Model Description, Current Prompt Questions, Model Predictions \& Targets, and Optimization Instruction:
	\begin{itemize}
		\item Instruction: The prompt template begins with an introduction outlining the responsibilities of the optimizer, clearly stating that its primary task is to optimize the guiding questions provided.
		\item Inputs: This section is used to attach the batched visual data for the reference of the optimizer.
		\item Model Description: The learning task of the learner is reiterated here for the information of the optimizer.
		\item Current Prompt Questions: The guiding questions used by the learner in the current iteration are shown here for the reference of the optimizer.
		\item Model Predictions \& Targets: The batched numerical predictions and the ground truths are shown here for $f_{\rm opt}$. These two inputs can tell the optimizer how well the learner does in the learning task on the mini-batch data.
		\item Optimization Instruction: The final section includes the instruction to ask the optimizer to think step by step with all the information above and output a new set of prompt questions with the required format.
	\end{itemize}

	\vspace{0.4mm}
	
	\begin{wrapfigure}{r}{0.3\linewidth}
		\centering
		\vspace{-5.5mm}
		\includegraphics[width=\linewidth]{ 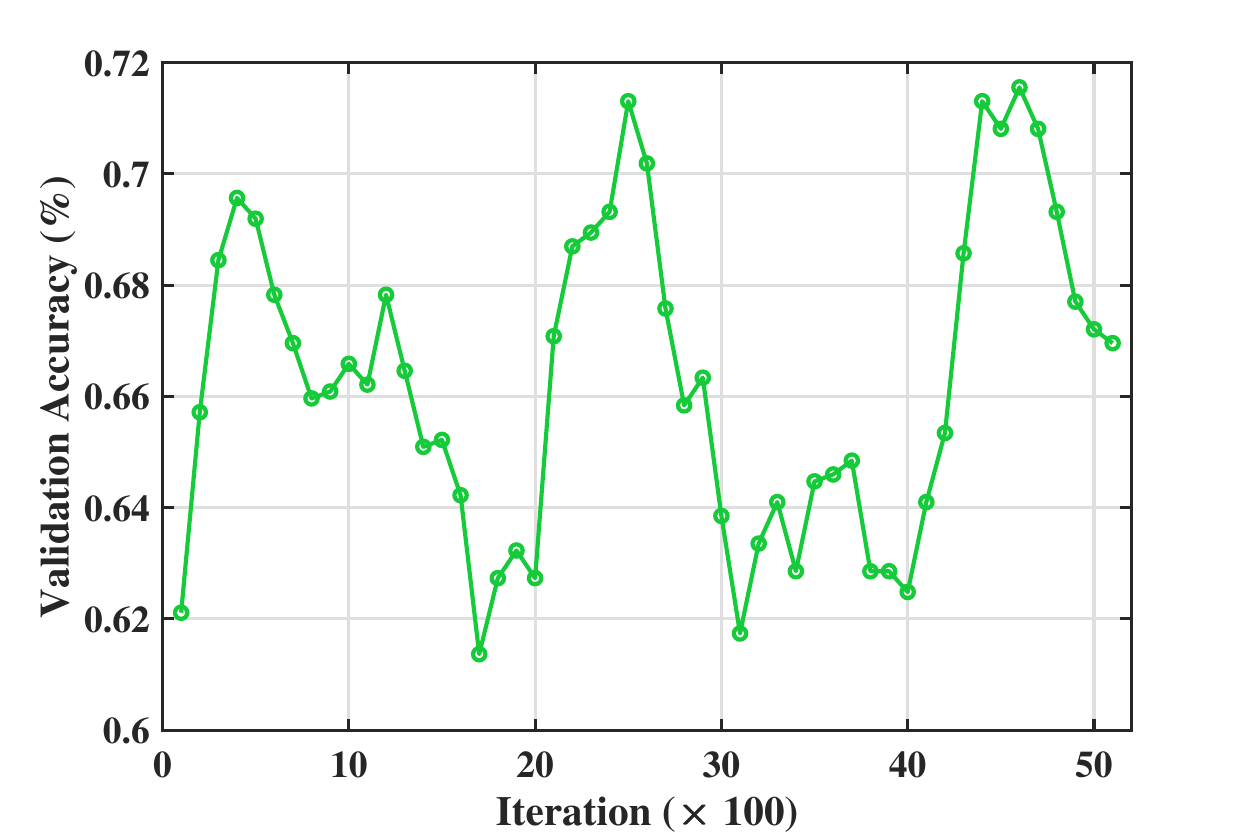}
		\vspace{-7mm}
		\caption{\footnotesize The validation accuracy given different learned guiding questions from each iteration. The graph is smoothed with moving average (window size 5) for better readability.}
		\label{fig:val_acc}
		\vspace{-2mm}
	\end{wrapfigure}

	\subsection{Details for Iterative Update by the Optimizer}
	\label{sec:optimizer}
	In training, we assess the quality of the learned guiding questions by the accuracy of the validation set. We show the validation accuracy from different questions $\mathbf{Q}_t$ obtained every 100 iterations (mini-batches) in Fig.~\ref{fig:val_acc}. In the duration of up to 5000 iterations in training, the observed plot in Fig.~\ref{fig:val_acc} contains three oscillations, each consisting of an increase in validation accuracy followed by a decrease. The increase represents that the optimizer VLM gradually finds better questions for the binary classification learning task when it sees more batched data, which shows the optimizer can understand its responsibility well and find better questions effectively. Meanwhile, we note that verbal optimization may not always lead to an increase. This is probably because the optimization is completely verbalized, and the VLM will have an inertial thinking behavior like humans, which gets the optimizer stuck in the wrong direction and makes it continue the optimization in a direction that is not beneficial. As a result, this causes the validation accuracy to decrease sometimes. Despite that, because of the guidance provided by the optimizer prompt template $\psi$, the optimizer can overcome its pitfalls in thinking and find good guiding questions in a new direction, which leads to an increase in validation accuracy afterward. This is an interesting phenomenon due to the distinction between verbal learning and traditional numerical optimization algorithms, and it will be a promising future direction to reduce the time in overcoming pitfalls in thinking for VLMs during VL. 
	
	\begin{figure*}[t]\centering\includegraphics[width=0.99\linewidth]{ 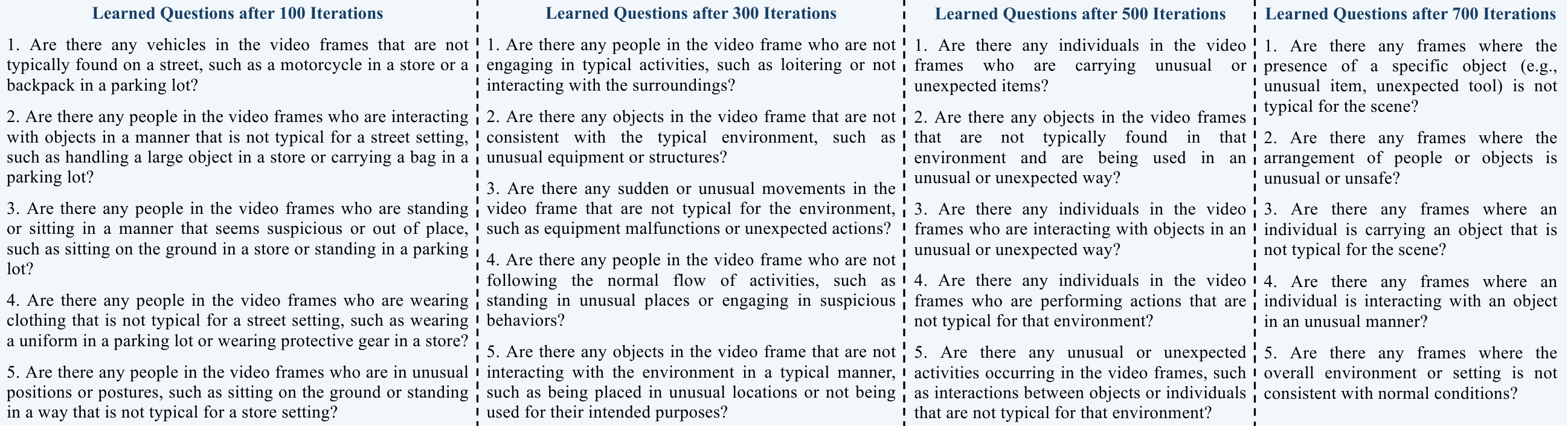}
		\vspace{-2mm}
		\caption{\footnotesize We take the  guiding questions $\mathbf{Q}$ learned from the 100th iteration to the 700th iteration for illustration purpose. During the updating process, the optimizer gradually concretizes anomaly patterns that can be applied to different scenarios in a concise expression. }
		\label{fig:q_show}
		\vspace{-2.5mm}
	\end{figure*}

	In addition, w.l.og., we take learned questions from the 100th iteration to the 700th iteration (which are within the first epoch) for illustration to show the process of updating $\mathbf{Q}$ by the optimizer in Fig.~\ref{fig:q_show}. First, as the optimizer sees more videos, it tries to make the questions focus on a more general setting. For example, the questions in the 100th iteration focus on ``street'' and ``store'' scenes. After more iterations, the questions become more generalizable for a general environment and focus on the elements that cause anomalies. Additionally, the anomalous pattern descriptions become more diverse as the optimization continues. To illustrate, in the beginning, the questions mostly pay attention to the humans, objects, and their interaction. In later iterations, the optimizers gradually summarize some previous questions into one and raise questions considering the overall environment (Q5 from the 700th iteration). Therefore, the VL framework proposed in this paper is effective in finding a diverse set of guiding questions for VAD that apply to general cases, which can elicit the reasoning of a frozen VLM in VAD.

	\newpage
	\section{Additional Experiments and Results}
	\label{sec:additional_exp}
	\subsection{Comparison to the State-of-the-art Methods on XD-Violence Measured by AP}
	\label{sec:ap}

	\setlength{\columnsep}{9pt}
	\begin{wraptable}{r}[0cm]{0pt}
		\hspace{-2.4mm}
		\setlength{\tabcolsep}{4.2pt}
		\renewcommand{\arraystretch}{1.09}
		\footnotesize
		\centering
		\begin{tabular}{ l|c}
			\specialrule{0em}{0pt}{-12pt}
			Method & AP\\
			\shline
			\multicolumn{2}{c}{\emph{Non-Explainable VAD Methods}}\\
			
			Wu et al.$^{\dagger}$~\cite{wu2020not}  &  78.64\\
			OVVAD~\cite{wu2024open} & 66.53 \\
			S3R$^{\dagger}$~\cite{wu2022self}& 80.26 \\
			RTFM$^{\dagger}$~\cite{tian2021weakly} & 77.81 \\
			MSL$^{\dagger}$~\cite{li2022self}   & 78.58 \\
			MGFN$^{\dagger}$~\cite{chen2023mgfn} &  80.11 \\
			CLIP-TSA$^{\dagger}$~\cite{joo2023clip}   & \textbf{82.19} \\
			\shline
			\multicolumn{2}{c}{\emph{Explainable VAD Methods}}\\
			Holmes-VAD$^{\dagger}$~\cite{zhang2024holmes}  & \textbf{84.96}\\
			LAVAD~\cite{zanella2024harnessing}  & 62.01\\
			ZS CLIP~\cite{zanella2024harnessing} &  17.83  \\
			{\scriptsize ZS IMAGEBIND-I}~\cite{zanella2024harnessing}  & 27.25  \\
			{\scriptsize ZS IMAGEBIND-V}~\cite{zanella2024harnessing}  & 25.36  \\
			LLAVA-1.5~\cite{liu2024improved}   &  50.26  \\\rowcolor{Gray}
			{\name} & 70.54  \\
			\specialrule{0em}{-7.25pt}{0pt}
		\end{tabular}
		\caption{\footnotesize AP~(\%) on XD-Violence. $\dagger$ indicates VAD methods are trained on entire training frames. No IT is used for Holmes-VAD.}
		\label{table:comparison_xd_ap}
		\vspace{-2mm}
	\end{wraptable}
	The comparison results regrading average precision (AP), \ie, the area under the frame-level precision-recall curve, on XD-Violence are shown in Table~\ref{table:comparison_xd_ap}. Compared to AUC, AP focuses on measuring the ability to identify the positive class (anomaly), while AUC measures how well a method separates anomaly and normalcy in general.  We provide the analysis of the results as follows.
	
	Firstly, under such a distinct property of AP, as pointed out by~\cite{wu2024open}, methods trained on the whole training set and utilizing all frames will enjoy advantages when measuring VAD performance by AP.  As a result, CLIP-TSA and Holmes-VAD, two methods using the whole training frames, attain the highest AP in the category of non-explainable and explainable VAD, respectively. We acknowledge there is a gap between {\name} and these two methods under AP on XD-Violence, which is understandable because they use the whole training frames to improve the ability to find anomalies of classifiers. To illustrate, in training {\name} only samples 8 frames for each video and only uses 0.19\% total frames (31,632 out of 16,378,527) for training on XD-Violence. Thus, our training is dramatically light compared to the methods like CLIP-TSA and Holmes-VAD in Table~\ref{table:comparison_xd_ap}. With fewer frames used for training, {\name} unavoidably achieve lower AP (which only considers positive cases) compared to those that have more, for it relies on fewer training data. In addition, we want to point out that judging the VAD performance solely by AP on XD-Violence can be biased. This is because the ratio of positive frames in XD-Violence (23.07\%) in test videos is overly higher than other datasets like UCF-Crime (7.92\%), which is unrealistic because the anomaly is sparse in the real world~\cite{sultani2018real}. Given that, only focusing on the comparison in AP on XD-Violence would amplify the bias in VAD performance evaluation, and we recommend taking into consideration other factors like training costs and the comprehensive ability of distinguishing anomaly and normality by the methods in evaluation.

	Secondly, among the methods (OVVAD, LAVAD, ZS CLIP, ZS IMAGEBIND, and LLAVA-1.5) that does not use full frames for training, {\name} achieves the best AP in this fair comparison, surpassing the second best method in the Explainable VAD category (LAVAD) over 8.53\%, which showcases the effectiveness of using learned guiding question to prompt frozen VLMs for VAD.

	To conclude, it is unfair to only judge VAD performance by AP on XD-Violence without considering the training costs and the relatively imbalanced frame distribution in test videos. Considering all factors into consideration, {\name} is a favorable method used for VAD in detecting anomalies.

	\subsection{Discussion on Generalizability of Used Questions}
	\label{sec:gen_vs_specific}
	During the optimization of $\mathbf{Q}$, because of the randomness involved in this process, the optimizer may output certain guiding questions that only focus on one specific surrounding. We find an interesting phenomenon on VLMs in VAD that guiding questions related to a specific scenario yield inferior VAD performance compared to the general questions in both general cases and specific cases.

	To illustrate, we take two sets of specific questions obtained on UCF-Crime for analysis. The first example is  a set of guiding questions $\mathbf{Q}_{\rm traffic}$ that only ask the VLM to consider anomalies related to the traffic as follows:
	\begin{enumerate}
		
		\item \emph{Are there any vehicles or people violating traffic rules?}
		\item \emph{Are there any accidents or near-accidents occurring?}
		\item  \emph{Are there any objects or people obstructing the normal flow of traffic?}
		\item \emph{Are there any unusual or unexpected behaviors from pedestrians or drivers?}
		\item \emph{Are there any emergency vehicles or personnel present?}
	\end{enumerate}
	
	The second example is another set of guiding questions $\mathbf{Q}_{\rm store}$ that only ask the VLM to identify anomalies in a store setting, which includes questions like:
	\begin{enumerate}
		\item   \emph{Are there any individuals loitering or behaving suspiciously inside the store? }
		\item  \emph{Is there any unusual activity inside the store, such as tampering with items or attempting to enter restricted areas?} 
		\item  \emph{Are there any signs of forced entry or damage to the store's entrance?} 
		\item  \emph{Are there any individuals present who seem to be watching or waiting for something specific inside the store?} 
		\item   \emph{Are there any interactions between individuals inside the store that appear suspicious or out of the ordinary?} 
	\end{enumerate}

	Thus, $\mathbf{Q}_{\rm traffic}$ and  $\mathbf{Q}_{\rm store}$ focuses on the specific anomalies of traffic accidents and shoplifting, respectively, while the $\mathbf{Q}^{*}$ that we find focuses on general cases and includes the following questions:
	\begin{enumerate}
		
		\item  \emph{Are there any people in the video who are not in their typical positions or engaging in activities that are not consistent with their usual behavior?}
		\item  \emph{Are there any vehicles in the video that are not in their typical positions or being used in a way that is not consistent with their usual function?}
		\item  \emph{Are there any objects in the video that are not in their typical positions or being used in a way that is not consistent with their usual function?}
		\item  \emph{Is there any visible damage or unusual movement in the video that indicates an anomaly?}
		\item  \emph{Are there any unusual sounds or noises in the video that suggest an anomaly?}
	\end{enumerate}

	The comparison results of $\mathbf{Q}^{*}$, $\mathbf{Q}_{\rm traffic}$, and $\mathbf{Q}_{\rm store}$ in detecting anomalies in general cases (all testing videos on UCF-Crime), traffic scenes (testing videos from the Traffic Accident category on UCF-Crime), and the store scenes (testing videos from the Shoplifting category on UCF-Crime) are shown in Table~\ref{tab:generalvsspecific}. It indicates that $\mathbf{Q}^{*}$ performs the best in both general cases and two specific cases like in traffic and store scenes.  This is because the overly specific definition of anomalies like  $\mathbf{Q}_{\rm traffic}$ and $\mathbf{Q}_{\rm store}$  makes it harder for a VLM to classify one clip into an anomaly and leads to more false negatives in its prediction given those specific questions, which degrades the performance. Therefore, we recommend using general questions like the ones shown in $\mathbf{Q}^{*}$ in frozen VLMs for VAD.

	\begin{table}[h]
		\centering
		
		\begin{tabular}{c|ccc}
			\specialrule{0em}{0pt}{2.5pt}
			\multirow{2}{*}{Questions} & \multicolumn{3}{c}{Scenario}\\
			& All & Traffic & Store \\
			\shline
			$\mathbf{Q}^{*}$ &  \textbf{86.55} & \textbf{70.43} & \textbf{72.58} \\
			$\mathbf{Q}_{\rm traffic}$ &  82.59&  
			67.53& / \\
			$\mathbf{Q}_{\rm store}$ & 76.67 &  
			/  & 44.84 \\
			
			\specialrule{0em}{-9pt}{0pt}
		\end{tabular}
		\caption{\footnotesize General guiding questions outperform specific ones measured by AUC (\%) on UCF-Crime. Specific questions are not tested on other specific scenarios, which is indicated by a slash (/).}
		\label{tab:generalvsspecific}
	\end{table}

	\subsection{Hyperparameters in Training}
	
	\noindent \textbf{Batch Size and Sampled Frame Number}. Key hyperparameters that need to be set in training are the batch size $n$  and the number of sampled frames $S$ for each video $V^{(j)}$ in the VL framework. 
	\setlength{\columnsep}{9pt}
	\begin{wraptable}{r}[0cm]{0pt}
		\setlength{\tabcolsep}{1pt}
		\renewcommand{\arraystretch}{1.1}
		\hspace{-2.4mm}
		\centering
		\footnotesize
		\begin{tabular}{c|c|c}
			\specialrule{0em}{0pt}{-8pt}
			
			Batch Size & Sampled Frames  & AUC (\%) \\
			\shline
			$n$ = 1 & $S$ = 16 &  81.53   \\
			$n$ = 2 &  $S$ = 8 &  \textbf{86.55} \\
			$n$ = 4 & $S$ = 4 &  83.19  \\
			$n$ = 8 & $S$ = 2 & 79.91  \\
			
			\specialrule{0em}{-7pt}{0pt}
		\end{tabular}
		\caption{\footnotesize The choice of batch size and sampling frames affects the effectiveness of the learned guiding questions in VAD. The results are obtained by InternVL2-8B as {\name}'s backbone.}
		\label{table:batch}
		\vspace{-2mm}
	\end{wraptable}
	They are correlated because they determine the total number of frames for the optimizer to skim and provide feedback as $S\cdot n$. Considering memory constraints when implementing VLMs on GPUs, we set $S\cdot n=16$ in training.  We further explore the trade-off between $S$ and $n$ given the constraints for input frames to decide $S$ and $n$. The results are shown in Table~\ref{table:batch}. If the batch size $n$ is 1 with $S=$ 16, the learned questions cannot be generalized due to the limited video sample in the batch which leads to a suboptimal AUC, and it takes longer to train for {\name}. Meanwhile, if we set $n$ as large numbers like 4 or 8 (with $S=$ 4 or $S=$ 2), the learned questions are suboptimal too because relatively few sampled frames generally lack the temporality for the optimizer to look into the details and conceive good questions. Thus, setting $n$ to 2 and $S$ to 8 is in default in this paper, which strikes the balance between training efficiency and effectiveness.

	\subsection{Hyperparameters in Inference and Sensitivity Test}
	\label{sec:hyperparams}

	\noindent \textbf{Hyperparameters in Inference}
	During inference, in Step 1, following~\cite{zanella2024harnessing}, the interval between each segment center $d$ is 16 frames. In Step 2, we use ImageBind~\cite{girdhar2023imagebind} as the feature extractor in computing segment similarity as ~\cite{zanella2024harnessing} does, and the number of retrieved segments $K$ depends on the total number of segments $h$ in each test video $V$. Setting $K$ to $(0.1\cdot h)$ to $(0.15\cdot h)$ is generally good. We set $K$ to $(0.1\cdot h)$ for UCF-Crime and to $(0.15\cdot h)$ for XD-Violence. The temperature $\tau$ in the Softmax function is set to 10 for both datasets in Eq.~\eqref{eq:weight}. In Step 3, due to the properties of datasets, we set the filter size $\omega$ of  $G(p)$ to 15 and $\sigma_1$ to 10  for UCF-Crime, while setting $\omega$ to 30 and $\sigma_1$ to 30 for XD-Violence. For position weighting, we set $c={\rm floor}(F/2)$  and $\sigma_2={\rm floor}(F/2)$ for both datasets to make sure the position weight covers the whole video sequence.

	W.l.o.g, we test the sensitivity of the VAD performance of {\name} regarding hyperparameters on UCF-Crime.

	\noindent\textbf{Sensitivity Test for $K$}. As shown in Table~\ref{table:K}, as the number of retrieved segments increases from 0 to $0.15\cdot h$, the AUC gradually increases from to 85.21\% to 86.61\%. Meanwhile, if we randomly select $0.1\cdot h$ segments for retrieval, the AUC is even lower than the performance without retrieval. Thus, using Eq.~\eqref{eq:weight} for retrieval is necessary. Meanwhile,  having a large $K$ greater than $0.15\cdot h$ will introduce some noise in Eq.~\eqref{eq:weight} and downgrade the AUC slightly. Thus, selecting $0.1\cdot h$ or $0.15\cdot h$ for $K$ is generally good choice.
	
	\begin{table}[h]
		\setlength{\tabcolsep}{10pt}
		\renewcommand{\arraystretch}{1.2}
		\centering
		\footnotesize
		\begin{tabular}{c|c|c|c|c|c|c}
			\specialrule{0em}{0pt}{-8pt}
			
			Ratio (\%) & 0 & 5& 10& 15 & 20 & 25   \\
			\shline
			AUC (\%) & 85.21 & 86.48& 86.55& 86.61& 86.42& 86.19\\
			
			\specialrule{0em}{-7pt}{0pt}
		\end{tabular}
		\caption{\footnotesize Influence of the number of retrieved segments on AUC. The AUC of not using retrieval (Ratio $=$ 0\%) and randomly selecting 10\% segments for Eq.~\eqref{eq:weight} is 85.21\% and 84.55\%, respectively.}
		\label{table:K}
		
	\end{table}

	\noindent\textbf{Sensitivity Test for $\omega$}. The filter size decides how many local segments are incorporated for the current segment for Gaussian smoothing. From Table~\ref{table: filter_size}, we find that AUC converges when the filter size increases to 15. Meanwhile, the VAD performance measured AUC is insensitive to $\omega$ and does not fluctuate much. Thus, we can set the filter size with a medium number like 15.
	
	\begin{table}[h]
		\setlength{\tabcolsep}{10pt}
		\renewcommand{\arraystretch}{1.2}
		\centering
		\footnotesize
		\begin{tabular}{c|c|c|c|c|c}
			\specialrule{0em}{0pt}{-8pt}
			
			$\omega$ & 5 & 10 & 15 & 20 & 25  \\
			\shline
			AUC (\%) & 86.25 & 86.43 &  86.55 & 86.61 & 86.60\\
			
			\specialrule{0em}{-7pt}{0pt}
		\end{tabular}
		\caption{\footnotesize Influence of filter size $\omega$ in Gaussian Smoothing on AUC.}
		\label{table: filter_size}
		\vspace{-2mm}
	\end{table}
	
	\noindent\textbf{Sensitivity Test for $\sigma_1$}. The AUC performance is also robust on the choice of $\sigma_1$. As, shown in Fig.~\ref{table: sigma_1}, when we set $\sigma_1$ greater than 1, the AUC generally remains around 86.50\%, which again shows the robustness of the design of anomaly scoring in {\name}. We can set $\sigma_1$ as 10 for {\name}.

	\begin{table}[h]
		\setlength{\tabcolsep}{10pt}
		\renewcommand{\arraystretch}{1.2}
		\centering
		\footnotesize
		\begin{tabular}{c|c|c|c|c|c}
			\specialrule{0em}{0pt}{-8pt}
			
			$\sigma_1$ & 1 & 5& 10 & 15 & 20   \\
			\shline
			AUC  (\%) & 86.17 & 86.49 &  86.55 & 86.49 & 86.54\\
			
			\specialrule{0em}{-7pt}{0pt}
		\end{tabular}
		\caption{\footnotesize Influence of $\sigma_1$ in Gaussian Smoothing on AUC.}
		\label{table: sigma_1}
		\vspace{-2mm}
	\end{table}

	\noindent\textbf{Sensitivity Test for $\tau$}.  The temperature hyperparameter $\tau$ in Eq.~\eqref{eq:weight} controls the entropy of the distribution obtained from the Softmax function while preserving the rank of each element.  As demonstrated in Table~\ref{table: tau}, when $\tau$ is a small number like 10e-8 that is close to 0, the distributions tend to become  a trivial distribution with all mass concentrated on the highest-probability class (corresponding to the segment itself), and the result is the same as the one by not using retrieval. As we gradually increase $\tau$ to a reasonably large number (from 0.01 to 1), the AUC value converges around 86.55\% with no obvious fluctuation, again proving the robustness of anomaly scoring in {\name} regarding hyperparameter selection. Note that when $\tau$ approaches $+\infty$, the distribution tends to become a uniform distribution, which yields an AUC of 86.59\%. From the discussion above, we  can generally choose $\tau$ to be an number in [0.01, 1] in implementation.
	
	\begin{table}[h]
		\setlength{\tabcolsep}{10pt}
		\renewcommand{\arraystretch}{1.2}
		\centering
		\footnotesize
		\begin{tabular}{c|c|c|c|c|c}
			\specialrule{0em}{0pt}{-8pt}
			
			$\tau$ & 10e-8 & 0.01& 0.1 & 1 & $+\infty$   \\
			\shline
			AUC  (\%) & 85.21 & 86.31 &  86.55 & 86.58 & 86.59\\
			
			\specialrule{0em}{-7pt}{0pt}
		\end{tabular}
		\caption{\footnotesize Influence of $\tau$ in Eq.~\eqref{eq:weight} on AUC.}
		\label{table: tau}
		\vspace{-2mm}
	\end{table}
	
	\noindent\textbf{Sensitivity Test for $\sigma_2$}. From Table~\ref{table: sigma_2}, we find that setting $\sigma_2=0.5F$ encodes the position information best in the anomaly score. A drop is noticeable if we choose $\sigma_2$ less than $0.5F$ for it will not cover the whole sequence, which is reasonable, while choosing a $\sigma_2$ great than $0.5F$ does not change much. Thus, based on the physical meaning of $\sigma_2$, which controls the width of the distribution, we should make $\sigma_2$ equal to $0.5F$ in anomaly scoring.

	\begin{table}[h]
		\setlength{\tabcolsep}{10pt}
		\renewcommand{\arraystretch}{1.2}
		\centering
		\footnotesize
		\begin{tabular}{c|c|c|c|c}
			\specialrule{0em}{0pt}{-8pt}
			
			$\sigma_2$ & w/o  Weighting & 0.25& 0.5 & 0.75    \\
			\shline
			AUC  (\%) & 85.48 & 85.43 &  86.55 & 86.27 \\
			
			\specialrule{0em}{-7pt}{0pt}
		\end{tabular}
		\caption{\footnotesize Influence of $\sigma_2$ in Position Weighting on AUC.}
		\label{table: sigma_2}
		\vspace{-2mm}
	\end{table}

	\begin{figure*}[t]
		\centering\includegraphics[width=0.99\linewidth]{ 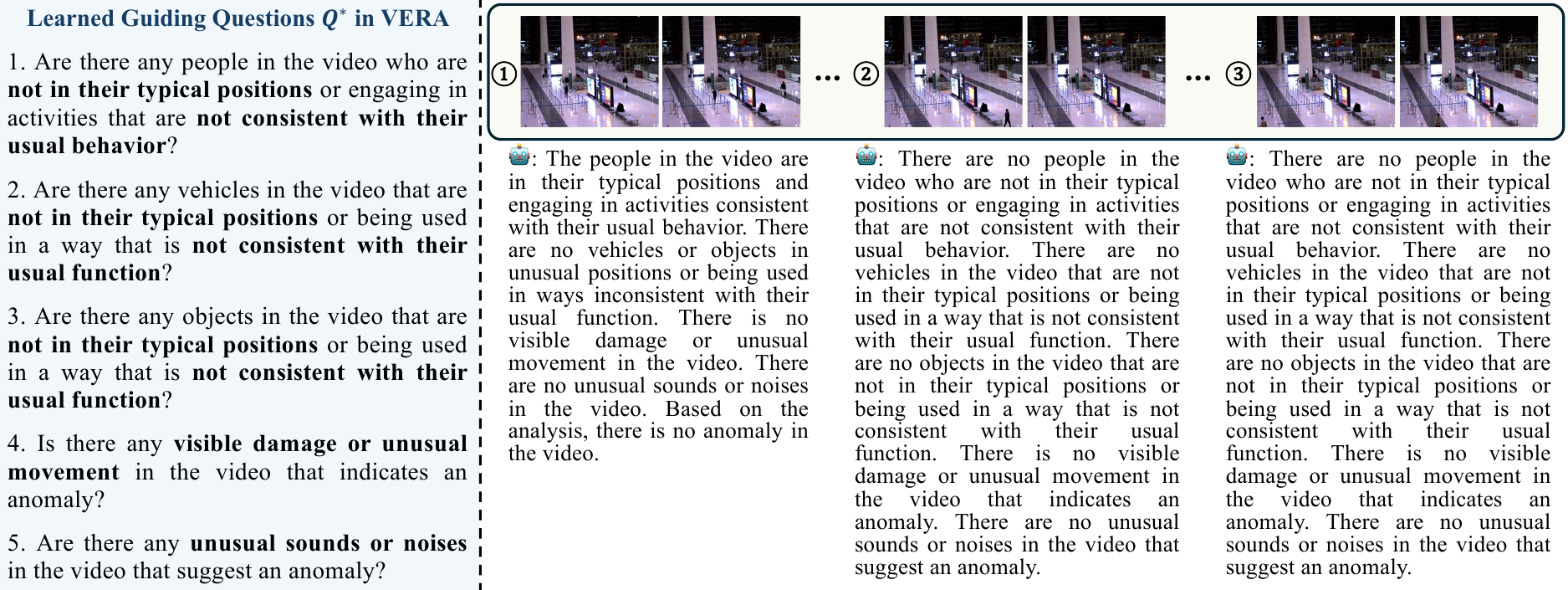}
		\vspace{-2mm}
		\caption{\footnotesize Given the normal video ``Normal\_Videos\_018\_x264'', the frozen VLM (InternVL2-8B) can conclude that no anomaly happens in the video  under the guidance of $\mathbf{Q}^{*}$, which is aligned with the ground truth. Since the anomaly scores for all scenes are zeros by {\name}, we do not show the complete anomaly scores with an additional figure.}
		\label{fig:case_normal}
		\vspace{-2.5mm}
	\end{figure*}

	\begin{figure*}[h]
		\centering\includegraphics[width=0.99\linewidth]{ 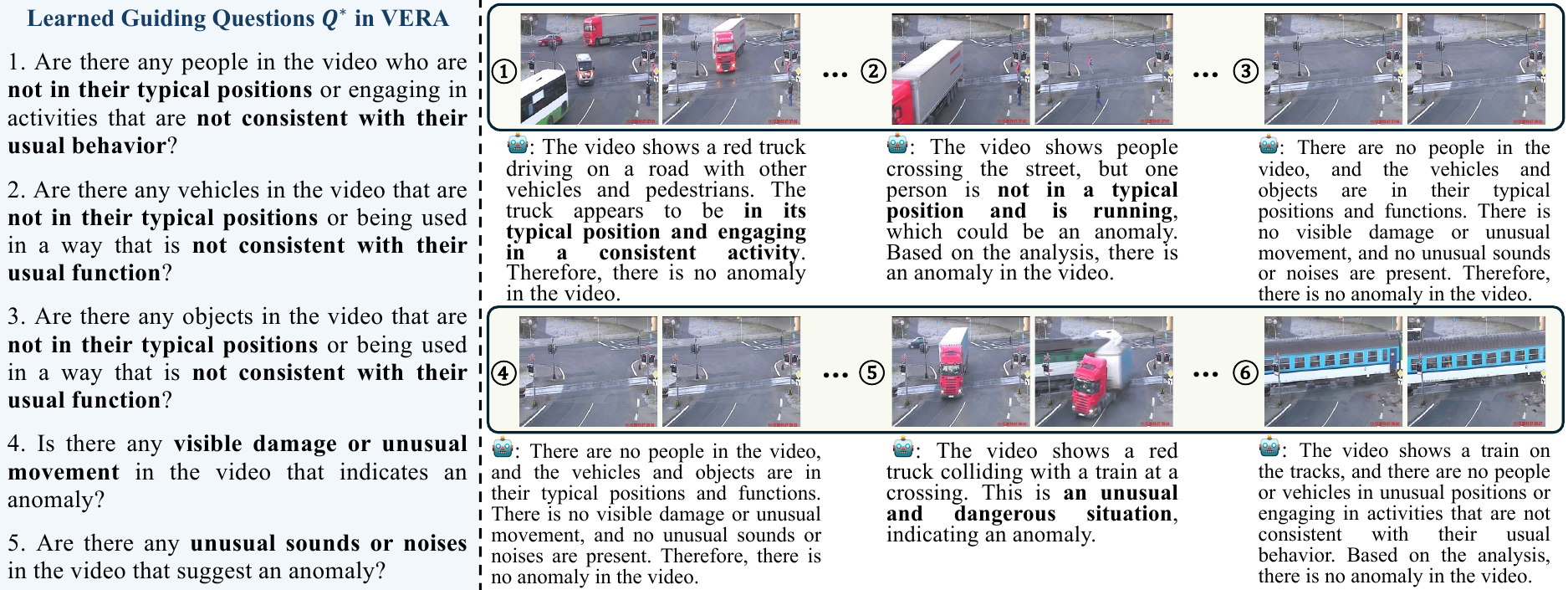}
		\vspace{-2mm}
		
		\caption{\footnotesize Given the abnormal video ``RoadAccidents127\_x264'', the frozen VLM (InternVL2-8B) can generate reasonable explanations aligned with the semantic change observed in each scene under the guidance of $\mathbf{Q}^{*}$. The complete anomaly scores are shown in Fig.~\ref{fig:score_case_3}. }    
		\label{fig:case_3}
		\vspace{-2.5mm}
	\end{figure*}

	\subsection{Additional Qualitative Results \& Case Studies}
	\label{sec:additional}
	W.l.o.g., we take one normal video (``Normal\_Videos\_018\_x264'')  and another abnormal video (``RoadAccidents127\_x264'') from the UCF-Crime dataset to demonstrate the explanations provided by a frozen VLM (InternVL2-8B) achieved by using the learned guiding questions $\mathbf{Q}^{*}$.

	First, in Fig.~\ref{fig:case_normal} we showcase the explanation of anomaly scoring by {\name} regarding a normal video ``Normal\_Videos\_018\_x264'' in UCF-Crime, which is taken in an airport hallway where no anomaly happens. For this video, {\name} assigns a 0 score to each frame. As shown in Fig.~\ref{fig:case_normal}, for the selected scenes in this video, {\name} explains that this is because there are no events that conform to the anomaly descriptions in $\mathbf{Q}^{*}$. Such explanations are consistent with the recording and again manifest the effectiveness of eliciting the reasoning ability in a frozen VLM for VAD by using learned guiding questions. Note that we do not have an additional figure illustrating the anomaly score dynamic for this video because all scenes are assigned 0 scores by {\name}. Next, we select 6 representative scenes in the abnormal video (``RoadAccidents127\_x264'') and show the corresponding explanation provided by the frozen VLM in Fig.~\ref{fig:case_3}. 
	\begin{wrapfigure}{r}{0.3\linewidth}
		\centering
		\vspace{-4.5mm}
		\includegraphics[width=\linewidth]{ 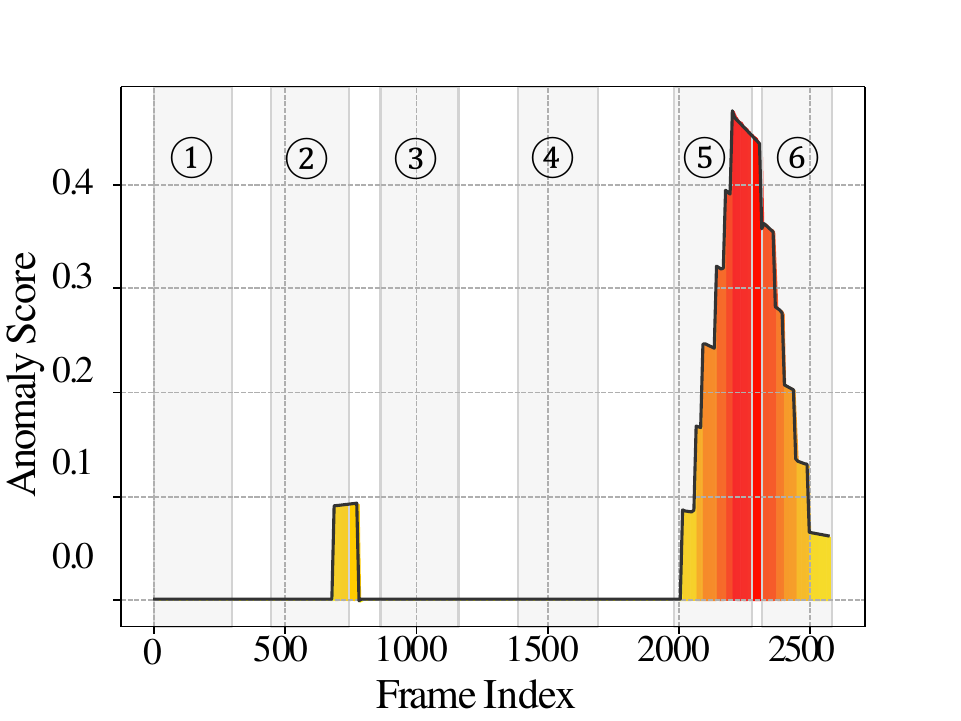}
		\vspace{-7mm}
		\caption{\footnotesize Anomaly scores generated by {\name} (with InternVL2-8B)  in ``RoadAccidents127\_x264'' from UCF-Crime.}
		\label{fig:score_case_3}
		\vspace{-2mm}
	\end{wrapfigure}
	The main anomaly that happens in this video is a traffic accident where a truck crashes into a train from Frame 2160 to Frame 2299, which corresponds to the 5th scene in Fig.~\ref{fig:case_3}. In particular, the figure shows that the learned question ``Is there any visible damage or unusual movement in the video that indicates an anomaly?'' in $\mathbf{Q}^{*}$ makes the frozen VLM find a good way to express what it sees in the 5th scene and understand this is an anomaly because the crash is unusual and dangerous. The other scenes are also well explained by the frozen VLM under $\mathbf{Q}^{*}$. Thus, this again verifies that the learned guiding questions can successfully trigger reasonable explanations in the adopted frozen VLM for VAD.

	Meanwhile, we also include the anomaly scores generated by {\name} for the abnormal video in Fig.~\ref{fig:score_case_3}. Most frames are assigned to zero except the scenes when someone crosses the road at an unusual speed (the 2nd scene in Fig.~\ref{fig:case_3}) and the truck-train crash happens (the 5th scene in Fig.~\ref{fig:case_3}). This fluctuation is aligned with the ground truth annotation and common sense about an anomaly, which shows that the anomaly scoring proposed in {\name} is reasonable.
	
	\section{Further Discussion on Limitations}
	\label{sec:limitation}
	Like existing VLM-based VAD methods, {\name}'s performance relies heavily on the visual perception capabilities of VLMs. Most VLMs employ the CLIP vision encoder~\cite{radford2021learning}, which has limitations in capturing fine-grained visual details. This limitation can impair precise anomaly detection. If important visual features are missing during the visual encoding process, then it is unlikely for {\name} to perform meaningful VL. 
	Therefore, a fundamental challenge for VLM-based VAD is to ensure sufficient visual and temporal features are encoded. Having verified this capability, {\name} can perform VL to extract crucial cues that guide video anomaly reasoning.

\end{document}